\documentclass[11pt]{article}

\PassOptionsToPackage{table}{xcolor}
\usepackage[final]{acl}
\usepackage{baostyle}
\usepackage{times}
\usepackage{latexsym}
\usepackage{enumitem}
\usepackage[T1]{fontenc}
\usepackage[utf8]{inputenc}
\usepackage{microtype}
\usepackage{inconsolata}
\usepackage{cleveref}
\usepackage{graphicx}
\usepackage[table]{xcolor}
\usepackage{tikz}
\usetikzlibrary{calc,positioning,fit,shapes.symbols,shapes.geometric}
\usepackage{booktabs}
\usepackage{tabularx}
\usepackage{pifont}
\usepackage{pdflscape}
\usepackage{ragged2e}
\usepackage{array}
\usepackage{longtable}
\newcommand{\cmark}{\ding{51}}% ✓
\newcommand{\xmark}{\ding{55}}% ✗
\definecolor{headergray}{HTML}{E0E0E0}
\definecolor{rowgray}{HTML}{F7F7F7}
\definecolor{donegreen}{HTML}{C8E6C9}
\definecolor{nextblue}{HTML}{BBDEFB}

\title{A Benchmark Construction and Evaluation Framework for Specialist Domains: Case Study on Defense-related Documents}
  \author{
  \normalfont
  Bao Gia Doan$^{\dagger}$,
  Aditya Joshi$^{\dagger}$,
  Pantelis Elinas$^{\dagger}$,
  Aarya Bodhankar$^{\dagger}$, \\
    Oscar Leslie$^{\ddagger}$, Tom Marchant$^{\ddagger}$,
    Flora Salim$^{\dagger}$\\
  $^{\dagger}$UNSW Sydney, Australia \\
  $^{\ddagger}$Cyndr AI, Australia \\
  \texttt{\{bao.doan1,aditya.joshi,p.elinas,a.bodhankar,flora.salim\}@unsw.edu.au} \\
  \texttt{\{oscar,tom\}@cyndr.ai}
}

\begin{document}
\maketitle
\begin{abstract}
RAG-based question-answering (QA) in specialist domains faces a cold-start problem: lack of evaluative benchmarks and absence of labeled data for post-training. We present \textbf{DoRA (Domain-oriented RAG Assessment)}, a novel benchmark construction and evaluation framework using only a small set of specialist domain documents. DoRA systematically generates synthetic QA training and evaluation datasets with auditable evidence across five domain-specific intents. To mitigate same-pipeline circularity, DoRA's training and test splits use \textit{different LLM families} (Claude Sonnet for training; GPT-4o for test) drawn from \textit{disjoint seed-document corpora}. Instantiated on 40 defense-related documents (written in English), DoRA yields $\sim$6.6K curated instances. Compared against 8 LLM baselines over a benchmark of 1,259 samples, a LoRA-adapted Llama3.1-8B trained on the synthetic training set consistently improves performance over 6 coverage and faithfulness metrics, especially reducing hallucination by more than half under the default GTE retrieval setting, with gains persisting across alternative retrievers and prompting-based baselines. Defense-domain expertise is incorporated in three stages of our evaluation: (a) determining the quality of the synthetic QA generated by DoRA, (b) ascertaining the reliability of LLM-as-judge scores, and (c) evaluating the generalization of the QA pipeline on completely human-written QA examples. We position DoRA as a practical framework for specialist-domain RAG under domain shift, with defense as a high-stakes case study.
\end{abstract}

\section{Introduction}
Retrieval-augmented generation (RAG) has rapidly become a popular paradigm for knowledge-intensive question-answering (QA) over \emph{domain-specific documents} \cite{lewis2020retrieval,DBLP:journals/corr/abs-2312-10997}. Yet RAG evaluation remains dominated by open-domain, Wikipedia-style benchmarks \cite{kwiatkowski2019natural,yang2018hotpotqa,joshi2017triviaqa,petroni-etal-2021-kilt}. In specialist domains such as legal, biomedical, defense, and regulated enterprise, three gaps recur: \textit{(i) absence of in-domain evaluation}, since no public benchmark exists for the target domain; \textit{(ii) mismatch of user intents}, since factoid QA under-represents the question types that practitioners ask~\cite{filice2025datamorgana,friel2024ragbenchexplainablebenchmarkretrievalaugmented}; and \textit{(iii) misaligned metrics}, since precision-oriented overlap scores (e.g., BLEU) reward surface similarity while users need attribution and faithfulness~\cite{zhu2024rageval}. The result is a cold-start problem: performance is hard to measure, supervised data is costly, and retrieval/generation iterations lack sufficient signal to localize the errors. We introduce DoRA (Domain-oriented RAG Assessment), a domain-grounded benchmark construction and evaluation framework that uses a small set of specialist-domain \textit{seed} documents to produce (a) a held-out benchmark split, (b) a larger synthetic split for supervised fine-tuning, and (c) a small expert-curated external check. Our design goals are: (1) \textbf{operability on small seed sets} (no reliance on Wikipedia-style coverage), (2) \textbf{practitioner-aligned question type coverage} (capture intents beyond factoid QA), and (3) \textbf{deployment-relevant, attribution-aware scoring} using modern RAG evaluation protocols \cite{es2023ragas,ming2024faitheval,chen2024benchmarking,friel2024ragbenchexplainablebenchmarkretrievalaugmented,zhu2024rageval}. In this paper, \textbf{we instantiate DoRA on publicly available Defense and National Security material from Australia}, using an Alpaca-inspired taxonomy of user intents. We treat Defense as a high-stakes case study that exercises the framework end-to-end, not as a claim that the benchmark is limited to defense alone. To the best of our knowledge, this is the first synthetic RAG benchmark and evaluation framework specifically instantiated on a defense and national security corpus---prior domain-specific synthetic QA work targets technical manuals~\cite{yuen2025automatic}, enterprise corpora~\cite{filice2025datamorgana}, or Wikipedia-style multi-hop QA~\cite{chen2023fewshotmultihop,abaskohi2025fmds,wu2024smmqg}. The contribution of DoRA is:

\begin{enumerate}[leftmargin=*, itemsep=0pt]
    \item \textbf{A novel domain-grounded benchmark construction and evaluation framework} for RAG-based QA where each QA is paired with auditable evidence bundles, enabling attribution-aware evaluation and reproducible benchmarking in specialized domains where expert time is costly\footnote{We will release publicly available document sources, prompts, reproducible code, and derived benchmark instances with provenance fields where document licenses and sensitivity constraints permit.}.
    \item \textbf{A lightweight validation protocol for specialized domains without large labeled test sets}. DoRA uses three low-cost checks: a retrieval-conditioned split, an expert-curated evaluation set, and human--LLM-judge calibration on a stratified sample.
    \item \textbf{A defense-domain case study with benchmark-driven adaptation} on 40 defense documents split into train/test corpora, producing $\sim$6.6K instances.
    A LoRA-adapted Llama3.1-8B
    trained on the cross-generator train split 
    improves RAGEval grounding and reduces hallucination roughly two-fold; the trend persists under alternative retrieval and expert-curated evaluation.
    \item \textbf{Manual evaluation by defense-domain experts} to evaluate dataset quality, LLM-as-judge reliability and generalization to a completely human-written dataset.
\end{enumerate}

\section{Related Work}
QA benchmarks \cite{petroni-etal-2021-kilt,kwiatkowski2019natural,joshi2017triviaqa,rajpurkar2016squad} are largely open-domain and Wikipedia-centric, while retrieval benchmarks such as BEIR \cite{thakur2021beir} and multi-hop QA testbeds \cite{tang2024multihop} target retriever generalization or compositional reasoning rather than deployment over a new private corpus. For RAG evaluation, recent frameworks including RAGAS \cite{es2023ragas}, ARES \cite{saad2023ares}, FaithEval \cite{ming2024faitheval}, LLM-in-RAG benchmarking \cite{chen2024benchmarking}, RAGBench \cite{friel2024ragbenchexplainablebenchmarkretrievalaugmented}, and RAGChecker \cite{ru2024ragchecker} provide useful diagnostics for faithfulness, attribution, and hallucination. Their common assumption, however, is that a suitable evaluation set already exists. Closer to DoRA are approaches that synthesize or structure evaluation for specific domains. RAGEval \cite{zhu2024rageval} uses scenario templates and synthetic articles for scenario-controlled testing; KIQA \cite{yuen2025automatic} generates questions from technical manuals; and DataMorgana \cite{filice2025datamorgana} synthesizes Q\&A over private enterprise corpora with user-defined question categories. DoRA is an enhancement over past work in its combination of three elements in one framework: synthetic benchmark construction from authentic seed documents, evidence-linked scoring for grounding and attribution, and a lightweight validation protocol with an expert-answered external check.

\begin{table}[t]
\centering
\scriptsize
\setlength{\tabcolsep}{2.8pt}
\renewcommand{\arraystretch}{1.1}
\newcolumntype{C}[1]{>{\centering\arraybackslash}p{#1}}
\newcolumntype{R}[1]{>{\RaggedRight\arraybackslash}p{#1}}

\begin{tabular}{@{} R{.46\columnwidth} C{.07\columnwidth} C{.07\columnwidth} C{.07\columnwidth} C{.08\columnwidth} C{.12\columnwidth} @{}}
\toprule
\textbf{Work} & \textbf{S} & \textbf{E} & \textbf{MM} & \textbf{Seed} & \textbf{Dom} \\
\midrule
Few-shot Multi-hop QA ~\citep{chen2023fewshotmultihop}
& \cmark & \xmark & \xmark & \cmark & Open \\
FM2DS ~\citep{abaskohi2025fmds}
& \cmark & \xmark & \cmark & \cmark & Open \\
SMMQG ~\citep{wu2024smmqg}
& \cmark & \xmark & \cmark & \cmark & Open \\
Prompting-based SDG ~\citep{schmidt-etal-2024-prompting-based}
& \cmark & \xmark & \xmark & \xmark & Open \\
RAGAS ~\citep{es2023ragas}
& \xmark & \cmark & \xmark & N/A & Open \\
ARES ~\citep{saad2023ares}
& \xmark & \cmark & \xmark & N/A & Open \\
RAGChecker ~\citep{ru2024ragchecker}
& \xmark & \cmark & \xmark & N/A & Open \\
RAGEval ~\citep{zhu2024rageval}
& \cmark & \cmark & \xmark & \xmark & Scenario \\
KIQA Gen. ~\citep{yuen2025automatic}
& \cmark & \xmark & \xmark & \cmark & Domain \\
DataMorgana ~\citep{filice2025datamorgana}
& \cmark & \xmark & \xmark & \cmark & Domain \\
\midrule
\textbf{\ours~(Ours)}
& \cmark & \cmark & \xmark & \cmark & \textbf{Domain} \\
\bottomrule
\end{tabular}

\caption{Comparison of synthetic QA generation and RAG evaluation.
S=Synthetic, E=RAG Evaluation, MM=Multimodal, Seed=Seed-doc grounded, Dom=Context scope (Open / Scenario / Domain).}
\label{tab:lit-compare}
\vspace{-5mm}
\end{table}

Synthetic pipelines such as DoRA reduce annotation costs and enable controlled stress tests. Related approaches include zero/few-shot prompting for QA generation \cite{ye-etal-2022-zerogen, schmidt-etal-2024-prompting-based, puri2020training}, few-shot multi-hop synthesis over Wikipedia pairs \cite{chen2023fewshotmultihop}, multimodal multi-hop synthesis \cite{abaskohi2025fmds}, and modality/style-conditioned question generation \cite{wu2024smmqg}. Compared to open-domain QA and retrieval benchmarks~\cite{kwiatkowski2019natural,yang2018hotpotqa,joshi2017triviaqa,petroni-etal-2021-kilt,thakur2021beir} and existing RAG evaluation frameworks \cite{es2023ragas,saad2023ares,ming2024faitheval,chen2024benchmarking,friel2024ragbenchexplainablebenchmarkretrievalaugmented,ru2024ragchecker}, DoRA enhances three aspects crucial to specialized domains: (i) grounding in authentic seed documents; (ii) coverage of five practitioner-aligned intent styles; and (iii) attribution-aware scoring paired with a small external validation protocol. A comparison with existing work is summarized in~\Cref{tab:lit-compare,tab:lit-compare-landscape}.

\section{Methodology}
\label{sec:method}

\begin{figure}[t]
    \centering
    \includegraphics[width=\linewidth]{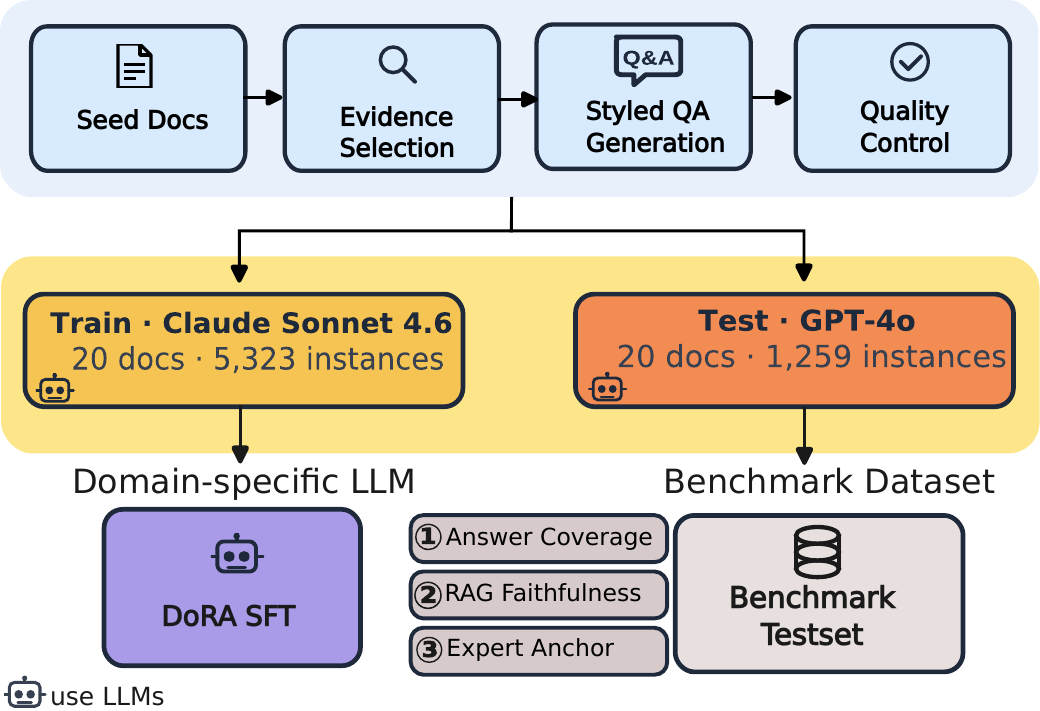}
    \caption{\ours pipeline, from seed documents to grounded-styled QA generation, and downstream domain evaluation and adaptation.}
    \label{fig:method}
    \vspace{-2mm}
\end{figure}
Let $\mathcal{C}$ denote a domain corpus segmented into passages with IDs, and let $\mathcal{S}$ be the five intent styles (\textsc{find}, \textsc{explain}, \textsc{summarize}, \textsc{generate}, \textsc{provide}). DoRA produces $\mathcal{D}=\{(q_i,a_i,s_i,E_i)\}_{i=1}^{N}$, where each instance is a quadruple $(q,a,s,E)$ comprising a question $q$, reference answer $a$, style $s \in \mathcal{S}$, and supporting passages $E \subset \mathcal{C}$. An LLM-based agent produces a prediction $\hat a$ for each question $q$. In RAG mode, we model this as:
$\hat a = G\!\left(q,\, R_k(q)\right)$ where $R_k$ is a retriever that returns the top-$k$ passages from $\mathcal{C}$ and $G$ is a generator that conditions on both the question and retrieved context (with the LLM-only setting as the special case $R_k(q)=\emptyset$). The pipeline of DoRA is in \Cref{fig:method}.

\subsection{Data Preparation}
The starting point is a \emph{minimal} input set: (i) a set of seed documents from the target domain, (ii) a short style specification (we use five intent styles prescribed by defense experts in this context: \textsc{find}, \textsc{explain}, \textsc{summarize}, \textsc{generate}, \textsc{provide}), and (iii) a bounded amount of expert time, reserved mainly for external validation rather than bulk labeling. In our case study, the seed set is 40 defense documents and the expert-time budget is small (\Cref{sec:human_annotated_dataset}). This matters because the framework is intended for settings where a large labeled corpus is unavailable and expert annotation time must be utilized strategically.

We construct DoRA from 40 \emph{seed documents} spanning defense strategy, capability, procurement, and industry-planning reports. 
Each document is chunked into passages, forming the corpus $\mathcal{C}$ used for evidence selection and retrieval.

\subsection{Synthetic Data Generation}
The dataset is created using three steps: first select a candidate evidence $E$ per seed, generate QA pairs conditioned on style and finally, perform quality control. For each style $s$, we select a small evidence set $E$ from the current seed and prompt a style-specific LLM template (see~\Cref{appd:prompts}) to generate $(q,a)$ conditioned on $(s,E)$; we store $E$ for auditability. Each style has an evidence budget $k_s$, which shapes evidence selection and downstream quality checks. We formulate the synthesis step as sampling from a \textit{style- and evidence-conditioned} generator: $(q,a)\sim p_{\phi}(\,\cdot \mid s,E\,)$, where $\phi$ denotes the parameters of the chosen LLM and prompt configuration (Chosen parameters in~\Cref{appd:detailed-method}).

\begin{table}[t]
\centering
\small
\setlength{\tabcolsep}{4pt}
\begin{tabular}{lrrrrr}
\toprule
\textbf{Style} & \textbf{\#} & \textbf{Avg $|q|$} & \textbf{Avg $|a|$} & \textbf{Refs} & \textbf{Qual.} \\
\midrule
\textsc{explain}   & 165 & 13.67 & 33.59 & 2.99 & 0.78 \\
\textsc{find}      & 421 & 12.65 & 20.83 & 1.00 & 0.88 \\
\textsc{generate}  & 309 & 13.76 & 31.42 & 3.34 & 0.78 \\
\textsc{provide}   & 155 & 14.00 & 30.45 & 1.91 & 0.76 \\
\textsc{summarize} & 209 & 12.85 & 29.35 & 2.65 & 0.82 \\
\midrule
\textbf{Total/Avg} & 1259 & 13.26 & 27.70 & 2.22 & 0.82 \\
\bottomrule
\end{tabular}
\caption{DoRA benchmark statistics by style: sample counts, average question/answer lengths (in tokens), average references per item, and average construction-time quality score.}
\label{tab:data-shape}
\vspace{-3mm}
\end{table}

\textit{(Task-conditioned) Evidence selection}: Evidence selection chooses a small evidence bundle (reference contexts) $E$ conditioned on the target style $s$, which determines both the desired cardinality $k_s \triangleq |E|$ and the selection behavior (e.g., extractive single-chunk evidence for \textsc{find} versus multi-chunk synthesis for long-form styles). During dataset construction, the candidate pool is typically restricted to passages from the current seed document, $\mathcal{C}_{\text{seed}}\subset \mathcal{C}$. We first apply a Natural Language Inference (NLI)-based prefilter to remove clearly irrelevant candidates and keep a reduced candidate pool $\widetilde{\mathcal{C}}_{\text{seed}}$ (top-$K_s$ by NLI relevance). We then use a local LLM scorer to rank candidate bundles drawn from $\widetilde{\mathcal{C}}_{\text{seed}}$. This scorer explicitly rates \emph{completeness}, \emph{complementarity} (non-redundant coverage), \emph{coherence}, and \emph{task fitness}; the highest-scoring bundle is returned. Formally, we select:

\begin{equation}
E^* \;=\; \arg\max_{\substack{E\subseteq \widetilde{\mathcal{C}}_{\text{seed}}\\ |E|=k_s}}\; g_s(E),
\end{equation}
where $g_s(E)$ is a style-conditioned \emph{bundle} scorer computed from the rubric dimensions above (with configurable weights and bounded search over candidate bundles). Details are in~\Cref{appd:detailed-method}.

\textit{Grounding and Quality Control}: Each instance stores an auditable evidence bundle $E$ (passage references), and we accept a generated item only if it satisfies \textit{style-specific} deployability constraints and meets an \textit{evidence-grounded} quality threshold:
\begin{equation*}
\begin{aligned}
\text{accept}(q,a,s,E)\;\Longleftrightarrow\;&\ h_s(q,a,E)=1 \\
&\wedge\ Q_s(q,a,E)\ge \tau_s,
\end{aligned}
\end{equation*}
where $h_s$ encodes style-specific hard checks (e.g., non-empty fields; \textsc{provide} must contain numeric content) and $Q_s$ is a composite, evidence-grounded score with threshold $\tau_s$.

We evaluate grounding with respect to $E$ using evidence-based metrics, including sentence-level natural language inference (NLI) with penalties for contradictions, extractive span matching (SQuAD-style span F1)~\cite{rajpurkar2016squad}, and checks for numerical and date consistency when applicable. For longer-form intents (namely, \textsc{explain}, \textsc{generate} and \textsc{summarize}), we additionally compute attribution-focused coverage (context recall and precision relative to $E$) and question--answer relevance (semantic similarity between $q$ and $a$), in line with standard RAG evaluation protocols (e.g.,~\citealp{es2023ragas}). Specifically, $Q_s$ combines a style-dependent set of metrics $\mathcal{M}_s$ using configurable weights $w_{s,m}$:
\begin{equation}
Q_s(q,a,E)=\sum_{m\in \mathcal{M}_s} w_{s,m}\, m(q,a,E),
\end{equation}
after which we apply deduplication and diversity filtering to eliminate repeated patterns and ensure balanced representation across the five styles. This procedure yields our \ours dataset, composed of the quadruples described above.

\subsection{Construction-time vs.\ benchmark-time evaluation} 

A run of DoRA on two \textit{disjoint} seed document sets yields two primary artifacts: (i) a held-out synthetic evaluation split for benchmarking and (ii) a cross-generator, cross-corpus training split for supervised fine-tuning. In addition, the same seed corpora for evaluation can support a separate set of expert-answered evaluations with modest effort from experts.
The quality controls described above are used only to \emph{construct} the benchmark by filtering candidate synthetic items. Once the dataset is fixed, benchmark-time evaluation is performed independently with \textit{held-out} questions, fixed retrievers or oracle evidence, and a separate three-layer metric stack (answer coverage, LLM-judge grounding diagnostics, and expert calibration; see~\Cref{sec:eval-setup}). This separation lets DoRA serve as a stable evaluation harness even as corpora, retrievers, prompts, and generators change over time.

\subsection{Note on cross-generator, cross-corpus construction}
\label{sec:cross-generator}
A standard pitfall in synthetic benchmarks is that the same LLM as well as source documents produce both the training supervision and the evaluation references, creating a train/test correlation that confounds genuine domain adaptation with pipeline-specific style mimicry. DoRA addresses this on two independent axes simultaneously: the training split is generated by \textbf{Anthropic Claude Sonnet} while the test split is generated by \textbf{OpenAI GPT-4o}, both conditioned on the same style-specific templates and quality controls (\textit{cross-generator separation}); and the two splits are drawn from \textit{disjoint} sets of seed documents with no document-level overlap, verified by passage-level n-gram and embedding-similarity checks (\textit{cross-corpus separation}). Construction-time judges are also drawn from a different model family (\textbf{Google Gemma}) than either the generators or the baselines to limit family-level coupling. Together, these choices decouple SFT supervision from the benchmark on the two axes most often cited as causes of inflated synthetic-evaluation results, so a model fine-tuned on the training split cannot win on the test split by imitating GPT-4o's writing style or by memorizing test-corpus content. One direct consequence is that train and test reference answers have systematically different length distributions (Claude-generated answers are longer), which shapes how token-overlap metrics should be interpreted; we return to this in the metric-design discussion of~\Cref{sec:eval-setup}.

\section{Experiment Setup}
\label{sec:eval-setup}
\paragraph{Data and splits.} We partition publicly available defense documents (written in English for a first-language user community) into two disjoint 20-document corpora: one used \emph{only} for training-split generation, the other \emph{only} for test-split generation. The held-out test split (1{,}259 instances, GPT-4o-generated; \Cref{tab:data-shape}) and the training split (5{,}323 instances, Claude Sonnet 4.6-generated, 90/10 train/eval; \Cref{appd:dataset}) are disjoint at both document and instance levels. We evaluate retriever performance (Hit@$k$, Recall@$k$; see \Cref{app:retrieval-metrics}) and end-to-end QA under two retriever conditions: a fixed dense retriever (default GTE-multilingual~\cite{alibaba2025gte}; IBM Granite~\cite{ibm2025graniteembeddingr2} as a robustness check) and an \emph{oracle} retriever that returns the stored gold evidence bundle $E$ to isolate generation from retrieval error.

\begin{figure}[t]
    \centering
    \includegraphics[width=.7\linewidth]{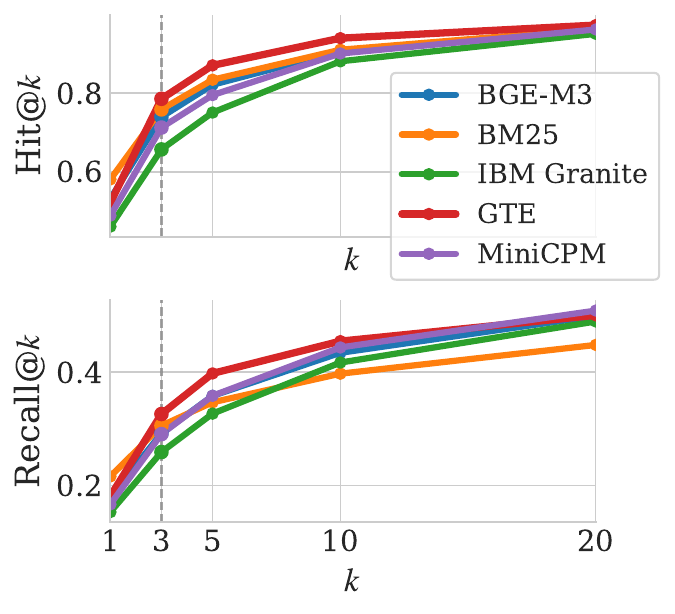}
    \caption{Retriever performance across top-$k$ on \ours test set.}
    \label{fig:retriever-topk}
    \vspace{-2mm}
\end{figure}

\paragraph{Why standard QA metrics are insufficient.} The cross-generator construction (\Cref{sec:cross-generator}) induces systematic length asymmetry between Claude-generated training references, GPT-4o-generated test answer references, and SFT-model outputs. In this regime, precision-oriented overlap scores (Token F1, BLEU, ROUGE-L F1) penalize verbose-but-correct answers, learned sentence-pair scores absorb length and reference-style cues, and \emph{none} of these reference-comparison metrics measure whether the answer is grounded in the retrieved evidence---the property a deployed RAG system is judged on.

\paragraph{Three-layer metric stack.} We, therefore, adopt a three-layer evaluation in which each layer compensates for a different weakness of the others. \textit{(i) Answer coverage}: recall-oriented variants of three established metrics spanning lexical to semantic granularity---Token Recall, ROUGE-L Recall~\cite{lin2004rouge}, and BERTScore Recall~\cite{zhang2019bertscore}---plus Length Ratio (generated/gold answer word-count) as a transparency control. All three coverage metrics are length-fair to verbose grounded answers because additional generated tokens do not reduce the score so long as the gold answer content is covered. \textit{(ii) Grounding at scale}: the RAGEval rubric judge~\cite{zhu2024rageval}, which extracts gold-answer keypoints and labels each as covered correctly (\textit{completeness}), addressed incorrectly (\textit{hallucination}), or unaddressed (\textit{irrelevance}); this is the only practical keypoint-level attribution signal at the 1{,}259-item benchmark scale (\Cref{app:rageval-metrics}). \textit{(iii) Expert ground-truth anchor}: a \textbf{50-item stratified human evaluation} (\Cref{sec:human-eval}) and \textbf{an independently authored 86-item expert-curated set} (\Cref{sec:human_annotated_dataset}), used to calibrate the LLM judge rather than scaled to the full benchmark. Full model identifiers, decoding settings, and per-style evidence budgets are listed in \Cref{tab:core-model-choices,tab:evidence-budgets}.

\section{Results}

\subsection{Retriever Evaluation}
\label{sec:retrieval}
Following RAGEval~\cite{zhu2024rageval}, we evaluate four closely related retrievers, along with IBM Granite retriever~\cite{ibm2025graniteembeddingr2}, across various top-$k$ values on the DoRA benchmark test set. \Cref{fig:retriever-topk} shows that, on defense documents, BM25 provides strong early hits but lags dense retrievers in broader coverage at larger $k$, motivating dense/hybrid retrieval as the natural default for deployment-facing RAG. Overall, GTE achieves the strongest balance across Hit/Recall$@k$ on the DoRA benchmark set, so we adopt it as the default retriever for end-to-end QA below.

  \begin{table*}[t]
    \centering
    \setlength{\tabcolsep}{3pt}
    \small
    \resizebox{.9\linewidth}{!}{%
    \begin{tabular}{lc ccc ccc}
    \toprule
    & & \multicolumn{3}{c}{\textbf{Answer Coverage (Recall)}} & \multicolumn{3}{c}{\textbf{RAG
  Faithfulness (LLM judge)}} \\
    \cmidrule(lr){3-5}\cmidrule(lr){6-8}
    Model & Len.~Ratio & ROUGE-L-R$\uparrow$ & Tok-R$\uparrow$ & BERTScore-R$\uparrow$ &
  Comp.$\uparrow$ & Hall.$\downarrow$ & Irrel.$\downarrow$ \\
    \midrule
    GPT-4o-mini                         & 3.52 & 66.99 & 67.29 & 88.94 & 67.81 & 4.92 &
  27.27 \\
    GPT-3.5-Turbo                       & 2.03 & 64.07 & 64.83 & \textbf{89.73} & 65.87 & \textit{3.71} &
  30.42 \\
    Llama-3.1-8B-Instruct (Base)        & 3.02 & 64.37 & 63.73 & 88.26 & 67.43 & 4.44 &
  28.13 \\
    Llama-3-8B-Instruct                 & 3.14 & 63.97 & 63.49 & 88.01 & 65.41 & 5.24 &
  29.35 \\
    GPT-4o                              & 2.47 & 65.96 & 66.10 & 89.07 & \textit{69.18} & 5.60 &
  25.22 \\
    Ministral-8B-Instruct               & 4.99 & 65.89 & 58.23 & 86.79 & 67.99 &  4.70 & 27.31 \\
    Claude-3.5-Haiku                    & 4.70 & \textit{68.02} & \textit{68.53} & 88.18 & 70.61 & 5.87 & \textbf{23.53} \\
    Qwen2.5-7B-Instruct                 & 3.21 & 66.14 & 66.45 & 89.20 & 66.98 & 4.65 &
  28.36 \\ \midrule

      \textbf{\ours{} SFT mean$\pm$std}   & {5.09{\scriptsize $\pm$0.08}} & \textbf{69.12{\scriptsize $\pm$0.19}} & \textbf{71.12{\scriptsize $\pm$0.16}} & \textit{89.11{\scriptsize $\pm$0.07}} & \textbf{73.37{\scriptsize $\pm$0.40}} & \textbf{2.17{\scriptsize $\pm$0.35}} & \textit{24.47{\scriptsize $\pm$0.52}} \\

    \textit{Gain vs Base}               &  & +4.75 & +7.39 & +0.85 & +5.94 & +2.27 & +3.66 \\
    \bottomrule
    \end{tabular}%
    }
    \caption{End-to-end QA with GTE retrieval (1{,}259 test samples, top-$k{=}3$). \textbf{Answer Coverage} reports recall-oriented variants spanning lexical to semantic granularity (Token Recall, ROUGE-L Recall, BERTScore Recall).
    \textbf{RAG Faithfulness} reports RAGEval keypoint-level completeness/hallucination/irrelevance under an LLM rubric judge. Len.~Ratio is generated/gold answer word-count ratio (transparency control). All metrics are \% except Len.~Ratio. 
    For \ours{} SFT, we report mean$\pm$std over three fine-tuning seeds. Gain vs Base is computed from the \ours{} SFT mean against Llama-3.1-8B-Instruct (base).
    \textbf{Bold}/\textit{italic} mark best/second-best. $\uparrow$ higher is better; $\downarrow$ lower is better.}
    \label{tab:qa-rageval-gte}
    \vspace{-2mm}
  \end{table*}

\subsection{End-to-End QA and DoRA SFT}
\label{sec:qa}
We evaluate end-to-end QA with a fixed retriever (GTE-multilingual-Base, top-$k$), where we use $k{=}3$ as a modest-context deployment setting that keeps the generator input budget controlled while still allowing multi-passage evidence to surface. Following the metric design in~\Cref{sec:eval-setup}, we report the recall-oriented coverage metrics (Token Recall, ROUGE-L Recall, BERTScore Recall) alongside the RAGEval grounding diagnostics; appendix tables (\Cref{appd:gold-context,appd:qa-ibm-granite}) report the same metric set under oracle and IBM Granite retrieval. 
The results in~\Cref{tab:qa-rageval-gte} show that, under the same GTE retrieved context, stronger and larger off-the-shelf LLMs remain competitive on individual semantic or irrelevance metrics, but no baseline dominates across the coverage and grounding diagnostics. This leaves room for domain adaptation: as detailed below, \ours{} SFT shifts the overall profile toward higher answer coverage and lower hallucination under the same retrieval and prompt setting.

\paragraph{DoRA SFT.} We Supervise-Fine-Tune (SFT) an open model (Llama3.1-8B-Instruct)\footnote{We choose Llama~3.1--8B for its deployment-friendly footprint and prior adoption in defense and national security AI contexts \cite{meta_llama31,cio_llama_defense}.} with LoRA~\cite{hu2021lora} on the cross-generator DoRA training split (5,323 Claude Sonnet 4.6 instances from the 20-document train-only corpus, 90/10 train/eval). Each example concatenates the question with evidence context $\mathrm{ctx}(E)$ and predicts the reference answer, mirroring end-to-end RAG. We optimize:
\begin{equation}
\min_{\theta}\sum_{(q,a,E)\in \mathcal{D}_{\text{train}}} -\log p_{\theta}\!\left(a \mid q, \mathrm{ctx}(E)\right),
\end{equation}
where $\theta$ denotes the (adapter) parameters. \Cref{tab:qa-rageval-gte} shows that the cross-generator-trained \ours{} SFT is stable across three fine-tuning seeds and improves over the base Llama-3.1-8B-Instruct most clearly on grounding: mean hallucination falls from 4.44\% to 2.17\%, with parallel gains in RAGEval Completeness and Irrelevance. Although \ours{} SFT does not top every metric (very large models like GPT-3.5-Turbo having the highest BERTScore Recall and Claude-3.5-Haiku getting the lowest Irrelevance), it gives the strongest combined coverage-grounding profile and improves over its own base on all reported metrics. This indicates stronger grounded keypoint coverage rather than only better reference overlap.
The Length Ratio shows that \ours{} SFT produces longer answers than LLM baselines, we treat length asymmetry as a model diagnostic and prioritize grounding and semantic diagnostics over raw overlap. Because the supervision comes from a different LLM family and disjoint document corpus, these gains are not well explained by same-family style matching or memorization of test passages. 

\begin{table}[t]
  \centering
  \small
  \setlength{\tabcolsep}{5pt}
  \renewcommand{\arraystretch}{1.05}
  \resizebox{\linewidth}{!}{%
  \begin{tabular}{lrrrr}
  \toprule
  \textbf{Setting} & \textbf{GPT-4o} & \textbf{Llama3.1} & \textbf{\ours{} SFT} & \textbf{Gain vs Base} \\
  \midrule
  GTE           & 69.18 & 67.43 & \textbf{73.37} & +5.94 \\
  IBM Granite   & 59.13 & 60.25 & \textbf{67.20} & +6.95 \\
  Gold evidence & 79.57 & 78.23 & \textbf{86.25} & +8.02 \\
  \bottomrule
  \end{tabular}
  }
  \caption{Downstream QA robustness across retrieval settings, measured by RAGEval Completeness (\%). 
  }
  \label{tab:qa-robustness}
  \vspace{-2mm}
  \end{table}

\subsection{Robustness across retrieval settings} 
The main trend persists across retrieval conditions (\Cref{tab:qa-robustness}), and the pattern of scores sharpens the deployment story. \ours{} SFT improves as retrieval quality improves: Completeness rises from 67.20\% with the weaker IBM Granite retriever to 73.37\% with the default GTE retriever. Its highest Completeness occurs under oracle evidence (86.25\%), where retrieval error is removed entirely, showing that adaptation improves not only robustness to imperfect retrieval but also the model's ability to use high-quality evidence. Full breakdowns are in~\Cref{appd:gold-context,appd:qa-ibm-granite}.

The cross-generator, cross-corpus construction (\Cref{sec:cross-generator}) already reduces same-family style alignment and document-level memorization risks. We now proceed to evaluate the performance based on \textit{whether or not evidence was retrieved}. We split the held-out test set by whether GTE retrieves any gold evidence chunk: Hit@$3{=}1$ (809; 64.3\%) vs.\ Hit@$3{=}0$ (450; 35.7\%). In \Cref{tab:qa-retrieval-conditioned}, \ours{} SFT is strongest when retrieval succeeds (84.38\% Completeness, 1.73\% hallucination), and its gain over the Llama-3.1-8B base grows when retrieval misses (+6.17 Completeness points, +4.65 hallucination points). Thus, the benefit of \ours{} SFT is not limited to easy retrieved contexts: it also improves robustness when retrieval fails to surface the gold evidence, suggesting that the adapted model can better use partial or noisy context instead of blindly following irrelevant passages.

\begin{table}[h]
  \centering
  \footnotesize
  \setlength{\tabcolsep}{2.5pt}
  \renewcommand{\arraystretch}{1.05}
  \resizebox{\columnwidth}{!}{%
  \begin{tabular}{lrrrrrr}
  \toprule
  \multirow{2}{*}{\textbf{Model}} & \multicolumn{3}{c}{\textbf{Hit@$3{=}1$}} & \multicolumn{3}{c}{\textbf{Hit@$3{=}0$}} \\
  \cmidrule(lr){2-4} \cmidrule(lr){5-7}
  & \textbf{Comp.}$\uparrow$ & \textbf{Hall.}$\downarrow$ & \textbf{Irr.}$\downarrow$
  & \textbf{Comp.}$\uparrow$ & \textbf{Hall.}$\downarrow$ & \textbf{Irr.}$\downarrow$ \\
  \midrule
	  GPT-4o          & 81.74 & 2.81 & 15.45 & 46.59 & 10.63 & \textbf{42.78} \\
	  Llama-3.1-8B-Inst & 79.21 & 2.96 & 17.83 & 46.25 &  7.11 & 46.64 \\ \midrule
	  \ours{} SFT     & \textbf{84.38} & \textbf{1.73} & \textbf{13.89} & \textbf{52.42} & \textbf{2.46} & \textit{45.12} \\
	  \textit{Gain vs Base} & +5.17 & +1.23 & +3.94 & +6.17 & +4.65 & +1.52 \\
	  \bottomrule
	  \end{tabular}%
	  }
	  \caption{Retrieval-conditioned QA under the main GTE setting. Hit@$3{=}1$ covers 809 samples (64.3\%); Hit@$3{=}0$ covers 450 samples (35.7\%). 
      }
  \label{tab:qa-retrieval-conditioned}
  \vspace{-2mm}
  \end{table}

\subsection{Domain Expert Evaluation}
\label{sec:human-eval}
We now report \textbf{human evaluation at three stages} as described below. Our human evaluators are defense domain experts who are not authors of the paper and are familiar with the national context of the documents.

\paragraph{Stage 1: Expert validation of synthetically generated QA pairs.} Four defense domain experts score a stratified 50-item benchmark sample with the same rubric used by our LLM judge. Inter-annotator Within-1 agreement is 94.7--100\% across the seven dimensions and $\kappa_w \ge 0.5$ on three of them (\Cref{tab:human-iaa}), and the LLM judge tracks experts substantially on groundedness ($\kappa_w{=}0.56$) and tightly within one point on relevance (96\% Within-1). This supports using the LLM judge as a scalable diagnostic for grounding-anchored signals, while keeping expert review for borderline items; full design and per-dimension results are in Stage-3 of this section.

\paragraph{Stage 2: QA Evaluation on Expert-curated QA pairs.} We create a separate expert-curated set of 86 samples whose reference answers are written and corrected by defense experts (without any use of LLMs), and are never shown to the evaluated LLMs (\Cref{sec:human_annotated_dataset}). On this harder held-out set, \Cref{tab:human-main} shows that \ours{} SFT roughly doubles Completeness (24.3\%) and halves hallucination (26.5\%) relative to the general-purpose baselines, with parallel gains in ROUGE-L and BERTScore Recall. The only weaker signal is Irrelevance: \ours{} SFT produces more content outside the gold keypoints. Because it also improves Completeness and reduces hallucination, we interpret this as verbosity, not worse factual grounding.

\begin{table}[h]
  \centering
  \scriptsize
  \setlength{\tabcolsep}{2pt}
  \renewcommand{\arraystretch}{1}
  \resizebox{\columnwidth}{!}{%
  \begin{tabular}{lccccc}
  \toprule
  \textbf{Model} & \textbf{Comp.}$\uparrow$ & \textbf{Hall.}$\downarrow$ &
  \textbf{Irr.}$\downarrow$ & \textbf{R-L-R}$\uparrow$ & \textbf{BERTS-R}$\uparrow$ \\
  \midrule
  GPT-4o          & 14.4          & 51.7          & \textbf{33.9} & 10.27          & 65.04
        \\
  Llama-3.1-8B    & 14.8          & 44.9          & 40.3          & 18.53          & 71.43
        \\ \midrule
  {\ours{} SFT}   & \textbf{24.3} & \textbf{26.5} & 49.2          & \textbf{39.20} &
  \textbf{80.94} \\

  \bottomrule
  \end{tabular}
  }
  \caption{Expert-curated evaluation. R-L-R=ROUGE-L Recall; BERTS-R=BERTScore Recall.}
  \label{tab:human-main}
  \vspace{-2mm}
  \end{table}

\begin{table}[t]
\centering
\small
\setlength{\tabcolsep}{4pt}
\renewcommand{\arraystretch}{1.05}
\begin{tabular}{lcccc}
\toprule
\textbf{Dimension} & \textbf{Exact} & \textbf{Within 1} & \textbf{MAD} & \textbf{$\kappa_w$} \\
\midrule
Comprehensiveness & 69.3\% & \phantom{0}97.0\% & 0.34 & \phantom{$-$}0.65 \\
Relevance         & 82.7\% & 100.0\% & 0.17 & \phantom{$-$}0.51 \\
Groundedness      & 85.3\% & \phantom{0}95.3\% & 0.19 & \phantom{$-$}0.48 \\
Verbosity         & 65.7\% & \phantom{0}94.7\% & 0.40 & \phantom{$-$}0.36 \\
Composition       & 87.0\% & \phantom{0}99.0\% & 0.14 & \phantom{$-$}0.29 \\
Clarity           & 87.7\% & \phantom{0}99.3\% & 0.13 & \phantom{$-$}0.55 \\
Alignment         & 96.3\% & 100.0\% & 0.04 & \phantom{$-$}0.16 \\
\bottomrule
\end{tabular}
\caption{Inter-annotator agreement among four domain-expert annotators on the seven-dimension rubric, computed over the 50 stratified benchmark items. 
}
\label{tab:human-iaa}
\vspace{-2mm}
\end{table}

\paragraph{Stage 3: Expert validation of LLM-as-Judge}
To calibrate our LLM-as-judge against expert judgment on the same benchmark items it scores, we conduct a human evaluation on a stratified sample of the DoRA test split. We draw 50 items balanced across the five intent styles (10 each), and four domain-expert annotators score every item independently 
using seven quality-dimension rubric on a 1--5 ordinal scale (\textit{comprehensiveness}, \textit{relevance}, \textit{groundedness}, \textit{verbosity}, \textit{composition}, \textit{clarity}, \textit{alignment}). Annotators see the (question, evidence, reference answer) triple, score in isolation without seeing the other annotators' labels, and use a fixed rubric definition document. \Cref{tab:human-iaa} shows strong expert consistency across the seven rubric dimensions: Within-1 agreement is 94.7--100\%, mean absolute difference is at most 0.40, and mean pairwise quadratic-weighted $\kappa_w$ is positive for all dimensions (0.16--0.65). The weakest $\kappa_w$ is on alignment, where expert scores saturate near the top of the scale, producing a ceiling effect rather than substantive disagreement.

We also calibrate the RAGEval LLM judge against expert scores on the three matching rubric dimensions: \textit{groundedness, comprehensiveness, and relevance}. On the 50-item sample, the judge aligns best on groundedness ($\kappa_w{=}0.56$, Within-1 92\%), is moderate on comprehensiveness ($\kappa_w{=}0.23$, Within-1 70\%), and is usually within one point on relevance (Within-1 96\%, MAD 0.19), where $\kappa_w$ is suppressed by expert-score saturation (\Cref{tab:human-judge-align}, \Cref{appd:human-judge-align}). These results support the three-layer framing of~\Cref{sec:eval-setup}: RAGEval grounding signals can be scored by the LLM judge at benchmark scale, while expert review serves as the ground-truth anchor for stylistic and coverage-sensitive dimensions.

\subsection{In-Context Learning Baseline}
\begin{figure}[h]
    \centering
    \includegraphics[width=\linewidth]{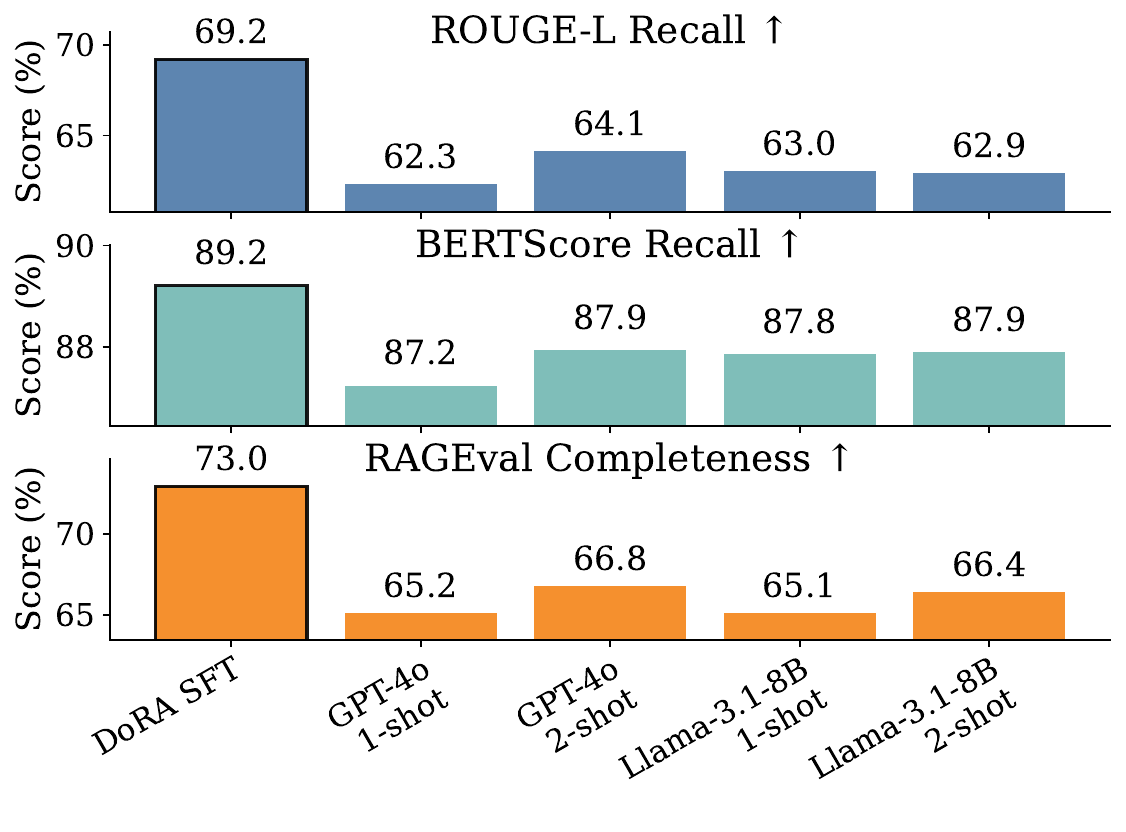}
    \caption{Answer Coverage (ROUGE-L Recall, BERTScore Recall) and RAGEval Completeness for \ours{} SFT vs.\ ICL baselines under GTE retrieval.}
    \label{fig:icl}
\end{figure}

As an adaptation-free baseline, we prompt GPT-4o and Llama-3.1-8B with 1-shot and 2-shot Q\&A exemplars curated from \ours{}, under the same GTE retrieval condition as~\Cref{tab:qa-rageval-gte}. \Cref{fig:icl} shows that ICL provides only marginal lift over the 1-shot setting---moving from 1-shot to 2-shot raises RAGEval Completeness by at most $\sim$2\,pp for both models (GPT-4o 65.2\%$\rightarrow$66.8\%, Llama-3.1-8B 65.1\%$\rightarrow$66.4\%). \ours{} SFT is clearly ahead at 73.0\% Completeness, with parallel gaps on ROUGE-L Recall and BERTScore Recall: curated exemplars provide useful format guidance, but fine-tuning on the cross-generator training split yields stronger grounded keypoint coverage.

\section{Discussion}
\label{sec:discussion}

\paragraph{Cross-generator design.} The cross-generator + cross-corpus construction (\Cref{sec:cross-generator}) is DoRA's most consequential design choice, and the same recipe transfers to any synthetic-benchmark plus SFT pipeline in a specialist domain at essentially no additional cost. DoRA, therefore, contributes a methodological archetype, not just a defense-domain artifact.

\paragraph{Tight clustering.} Under the same retriever configuration, off-the-shelf LMs cluster closely on semantic and grounding diagnostics (\Cref{tab:qa-rageval-gte}), suggesting that evaluation in private, domain-specific settings is bottlenecked by evidence selection and grounding rather than by raw model size. Targeted domain adaptation (even at 8B scale, via LoRA on cross-generator supervision) closes meaningfully more of the visible headroom than swapping in a larger general-purpose model. Notably, this benefit becomes more pronounced as retrieval quality degrades (\Cref{tab:qa-robustness}), matching deployment conditions where retrievers are imperfect.

\paragraph{From benchmark to deployment.} DoRA is designed for cold-start specialist settings and should be treated as an in-domain regression signal, not a universal leaderboard: teams rerun it after updating the corpus, retriever, prompts, or adaptation, and inspect where failures move across styles and grounding diagnostics. The construction recipe depends only on partitionable seed documents, an intent taxonomy, and two LLM families.

\section{Conclusion}

We presented DoRA, a domain-grounded benchmark construction and evaluation framework for cold-start RAG in specialist domains. DoRA starts from a small set of seed documents and produces auditable QA instances with linked evidence, enabling evaluation when no in-domain benchmark or large labeled dataset exists. Its cross-generator, cross-corpus design uses different LLM families and disjoint seed-document corpora for training and test generation, which structurally decouples SFT supervision from the benchmark and reduces the main circularity risks in synthetic evaluation. In a defense-domain case study, DoRA produces a held-out synthetic test split, a cross-generator SFT-ready training split, and an expert-curated external check. LoRA adaptation of Llama-3.1-8B substantially improves grounding diagnostics under the default GTE setting and remains robust across alternative retrieval conditions, showing that synthetic supervision can improve deployment-facing RAG behavior even under domain shift.

\clearpage
\section*{Limitations}
DoRA's empirical evaluation in this paper covers a single specialist domain (defense), so transfer of the framework to other specialist corpora is argued rather than demonstrated. Although the training and test splits are generated by different LLM families and drawn from disjoint seed documents to limit same-family style alignment and document-level memorization, they still share an overall construction pipeline (same prompts, same NLI prefilter, same quality controls), so within-pipeline metrics can still partially reflect pipeline-level style correlation; we mitigate this with retrieval-conditioned analysis and an independently constructed expert-curated check set, but cannot fully rule out residual correlation. The expert-curated check set is small (86 items), so statistical significance on this split is limited; we therefore use it as a qualitative external signal rather than a leaderboard. The framework depends on capable LLMs at multiple stages---generation, evidence scoring, and parts of evaluation---which can propagate shared biases or blind spots. Reported SFT numbers are averaged over three fine-tuning seeds, but we still report point estimates for most non-SFT baselines rather than full confidence intervals. Finally, DoRA measures benchmark quality and grounding, but does not measure pretraining contamination directly; teams adopting it in domains with public-document overlap should add a contamination probe.

\section*{Ethics Statement}
The case-study corpus is sourced from open-public defense documents rather than classified or proprietary material. We will release the seed documents, prompts, reproducible code, and derived benchmark instances with provenance fields, subject to document licenses and sensitivity constraints. The contribution of this work lies in the synthetic creation of a QA benchmark, an evaluation framework, and an adaptation setting. The defense application described in this paper is restricted to answering questions based on a fixed set of defense documents. The system does not make decisions, take actions, or replace human judgment; all interpretation and downstream use of its outputs remain the responsibility of human operators and fall outside the scope of this work. We expect that practitioners applying DoRA to other specialist domains will perform their own ethics review appropriate to that domain.

\bibliography{custom}

\appendix

\section*{Appendix}
\label{appd:overview}
This appendix provides additional results and implementation details for reproducibility:
\begin{itemize}
  \item \textbf{Additional QA Results} (\Cref{appd:additional-results}): QA with oracle-context (gold evidence)~(\Cref{appd:gold-context,tab:qa-rageval-oracle}), and end-to-end QA with IBM Granite retrieval (\Cref{appd:qa-ibm-granite,tab:qa-rageval-granite,tab:qa-retrieval-conditioned-granite}).
	  \item \textbf{Manually Curated Dataset} (\Cref{sec:human_annotated_dataset}): expert-answered set creation and its use as an external evaluation check.
	  \item \textbf{Human--LLM-Judge Alignment on RAGEval-Mapped Dimensions} (\Cref{appd:human-judge-align,tab:human-judge-align}): per-dimension calibration of the LLM judge against the mean expert score on the rubric dimensions that map to the RAGEval signals used in the main results.
  \item \textbf{Metrics} (\Cref{app:retrieval-metrics}): retrieval metrics (Hit@k, Recall@k), retriever identifiers (\Cref{tab:retriever-ids}), and implementation identifiers for learned metrics and judges (\Cref{tab:core-model-choices}).
	  \item \textbf{RAGEval Metrics} (\Cref{app:rageval-metrics}): definitions and formulas for the RAGEval diagnostics (completeness, hallucination, irrelevance) used in our end-to-end QA evaluation.
	  \item \textbf{\ours Dataset} (\Cref{appd:dataset}): dataset composition and generation statistics by style.
	  \item \textbf{Style Examples} (\Cref{appd:examples}): example questions and answers for each of the five intent styles.
	  \item \textbf{Detailed Methodology} (\Cref{appd:detailed-method}): model choices (\Cref{tab:core-model-choices}) and evidence-selection budgets (\Cref{tab:evidence-budgets}).
	  \item \textbf{Prompts} (\Cref{appd:prompts}): prompt templates used for benchmark-time grounded inference and style-conditioned QA generation.
	\end{itemize}

\section{Additional QA Results}
\label{appd:additional-results}

\subsection{QA with Gold Evidence Context}
\label{appd:gold-context}

To isolate generation quality from retrieval errors, we also report \emph{gold evidence context} results where the generator is given the recorded gold evidence passages $E$ (oracle context) instead of retrieved passages. \Cref{tab:qa-rageval-oracle} reports oracle-context results with the same prompt used across models. Under oracle evidence, grounding improves substantially across LLM models; \ours{} SFT gives the strongest lexical recall and lowest hallucination rate, while remaining competitive on Completeness and Irrelevance. This indicates that the adaptation benefit is not only a retrieval artifact.

\begin{table*}[t]
  \centering
  \setlength{\tabcolsep}{3pt}
  \small
  \resizebox{.8\linewidth}{!}{%
  \begin{tabular}{lc ccc ccc}
  \toprule
  & & \multicolumn{3}{c}{\textbf{Answer Coverage (Recall)}} & \multicolumn{3}{c}{\textbf{RAG Faithfulness (LLM judge)}} \\
  \cmidrule(lr){3-5}\cmidrule(lr){6-8}
  Model & Len.~Ratio & ROUGE-L-R$\uparrow$ & Tok-R$\uparrow$ & BERTScore-R$\uparrow$ & Comp.$\uparrow$ & Hall.$\downarrow$ & Irrel.$\downarrow$ \\
  \midrule
  GPT-4o-mini                         & 2.43 & 75.31 & \textit{74.35} & 92.03 & 83.20 &  3.40 & 13.40 \\
  GPT-3.5-Turbo                       & 1.61 & 71.88 & 71.79 & \textbf{92.11} & 78.99 &  \textit{2.69} & 18.32 \\
  Llama-3.1-8B-Instruct (base)        & 1.98 & 68.22 & 65.32 & 89.81 & 78.23 &  4.67 & 17.10 \\
  Llama-3-8B-Instruct                 & 2.08 & 69.47 & 66.80 & 90.03 & 77.52 &  4.20 & 18.28 \\
  GPT-4o                              & 1.66 & 72.41 & 71.51 & 91.31 & 79.57 &  4.23 & 16.21 \\
    Ministral-8B-Instruct               & 3.84 & 70.99 & 59.34 & 88.59 & 83.69 &  3.93 & {12.38} \\
    Claude-3.5-Haiku                    & 3.45 & \textit{76.23} & 74.09 & 90.91 & \textit{85.64} &  3.93 & \textbf{10.42} \\
  Qwen2.5-7B-Instruct                 & 2.27 & 74.25 & 72.87 & \textit{92.05} & 81.18 &  3.53 & 15.29 \\ \midrule
    \textbf{\ours{} SFT mean$\pm$std}   & {4.17{\small±0.08}} & \textbf{77.61{\small±0.38}} & \textbf{78.28{\small±0.40}} & {91.68{\small±0.03}} & \textbf{86.25{\small±0.76}} & \textbf{1.96{\small±0.04}} & \textit{11.79{\small±0.76}} \\

  \textit{Gain vs Base}               &  & +9.39 & +12.96 & +1.87 & +8.02 & +2.71 & +5.31 \\
  \bottomrule
  \end{tabular}%
  }
  \caption{End-to-end QA with oracle evidence (1{,}259 test samples). Len.~Ratio is generated/gold-answer word-count ratio; ROUGE-L, Tok-R, and BERTScore are recall variants; Comp./Hall./Irrel.\ are RAGEval diagnostics. All metrics are \% except Len.~Ratio. For \ours{} SFT, we report mean$\pm$std over three seeds. Gain vs Base is computed from the \ours{} SFT mean against Llama-3.1-8B-Instruct (base).}
  \label{tab:qa-rageval-oracle}
\end{table*}

\subsection{End-to-End QA with IBM Granite Retriever}
\label{appd:qa-ibm-granite}

We also report end-to-end QA under an alternative fixed retriever, IBM Granite (english-r2), to stress-test robustness to retrieval variation.

\begin{table*}[t]
  \centering
  \setlength{\tabcolsep}{3pt}
  \small
  \resizebox{.8\linewidth}{!}{%
  \begin{tabular}{lc ccc ccc}
  \toprule
  & & \multicolumn{3}{c}{\textbf{Answer Coverage (Recall)}} & \multicolumn{3}{c}{\textbf{RAG Faithfulness (LLM judge)}} \\
  \cmidrule(lr){3-5}\cmidrule(lr){6-8}
  Model & Len.~Ratio & ROUGE-L-R$\uparrow$ & Tok-R$\uparrow$ & BERTScore-R$\uparrow$ & Comp.$\uparrow$ & Hall.$\downarrow$ & Irrel.$\downarrow$ \\
  \midrule
  GPT-4o-mini                         & 3.16 & 62.15 & 62.35 & 87.22 & 61.64 &  7.85 & 30.51 \\
  GPT-3.5-Turbo                       & 1.70 & 55.49 & 56.29 & 86.58 & 57.36 &  8.20 & 34.44 \\
  Llama-3.1-8B-Instruct (base)        & 3.17 & 62.73 & 62.07 & 87.55 & 60.25 &  7.21 & 32.54 \\
  Llama-3-8B-Instruct                 & 3.07 & 61.25 & 60.63 & 87.26 & 59.34 &  6.57 & 34.08 \\
  GPT-4o                              & 2.04 & 56.53 & 56.28 & 85.32 & 59.13 &  9.27 & 31.60 \\
    Ministral-8B-Instruct               & 4.96 & 63.15 & 56.25 & 86.06 & 62.58 &  \textit{5.66} & 31.76 \\
    Claude-3.5-Haiku                    & 4.51 & \textit{65.00} & \textit{65.59} & 87.64 & \textit{65.18} &  6.94 & \textbf{27.88} \\
  Qwen2.5-7B-Instruct                 & 3.34 & 63.01 & 62.98 & \textit{87.94} & 62.20 &  6.01 & 31.79 \\ \midrule
    \textbf{\ours{} SFT mean$\pm$std}   & {5.11{\small±0.07}} & \textbf{66.47{\small±0.11}} & \textbf{68.58{\small±0.03}} & \textbf{88.40{\small±0.05}} & \textbf{67.20{\small±0.25}} & \textbf{3.44{\small±0.17}} & \textit{29.36{\small±0.35}} \\

  \textit{Gain vs Base}               &  & +3.74 & +6.51 & +0.85 & +6.95 & +3.77 & +3.18 \\
  \bottomrule
  \end{tabular}%
  }
  \caption{End-to-end QA with IBM Granite retrieval (1{,}259 test samples, top-$k{=}3$). Len.~Ratio is generated/gold-answer word-count ratio; ROUGE-L, Tok-R, and BERTScore are recall variants; Comp./Hall./Irrel.\ are RAGEval diagnostics. All metrics are \% except Len.~Ratio. For \ours{} SFT, we report mean$\pm$std over three seeds. Gain vs Base is computed from the \ours{} SFT mean against Llama-3.1-8B-Instruct (base).}
  \label{tab:qa-rageval-granite}
\end{table*}

\begin{table}[t]
  \centering
  \footnotesize
  \setlength{\tabcolsep}{2.5pt}
  \renewcommand{\arraystretch}{1.05}
  \resizebox{\columnwidth}{!}{%
  \begin{tabular}{lrrrrrr}
  \toprule
  \multirow{2}{*}{\textbf{Model}} & \multicolumn{3}{c}{\textbf{Hit@$3{=}1$}} & \multicolumn{3}{c}{\textbf{Hit@$3{=}0$}} \\
  \cmidrule(lr){2-4} \cmidrule(lr){5-7}
  & \textbf{Comp.}$\uparrow$ & \textbf{Hall.}$\downarrow$ & \textbf{Irr.}$\downarrow$
  & \textbf{Comp.}$\uparrow$ & \textbf{Hall.}$\downarrow$ & \textbf{Irr.}$\downarrow$ \\
	  \midrule
	  GPT-4o            & 80.64 & 4.26 & 15.11 & 32.02 & 15.59 & 52.39 \\
	  Llama-3.1-8B-Inst & 79.93 & 4.82 & 15.25 & 35.46 & 10.21 & 54.33 \\ \midrule
	  \ours{} SFT       & \textbf{84.02} & \textbf{1.95} & \textbf{14.02} & \textbf{45.74} & \textbf{4.76} & \textbf{49.51} \\
	  \textit{Gain vs Base} & +4.09 & +2.87 & +1.23 & +10.28 & +5.45 & +4.82 \\
	  \bottomrule
	  \end{tabular}%
	  }
	  \caption{Retrieval-conditioned QA under IBM Granite. Hit@$3{=}1$ covers 702 samples (55.8\%); Hit@$3{=}0$ covers 557 samples (44.2\%). Comp./Hall./Irr.\ are RAGEval diagnostics (\%). Gain vs Base is \ours{} SFT improvement over Llama-3.1-8B-Inst.}
  \label{tab:qa-retrieval-conditioned-granite}
  \end{table}

In this setting, where IBM Granite is weaker than GTE on retrieval metrics (\Cref{fig:retriever-topk}), \ours{} SFT \textit{widens its margin on grounding signals}: it achieves the lowest hallucination (3.44\%) and second-lowest irrelevance (29.36\%) in the aggregate table, more than halving hallucination relative to the base Llama-3.1-8B-Instruct (7.21\%). The retrieval-conditioned split in~\Cref{tab:qa-retrieval-conditioned-granite} shows where the gain comes from. Under successful Granite retrieval, \ours{} SFT has the highest Completeness (84.02\%) and lowest hallucination (1.95\%), improving over the base by +4.09 Completeness points and reducing hallucination by 2.87 points. Under retrieval failure, the margin grows: \ours{} SFT reaches 45.74\% Completeness, compared with 35.46\% for the base and 32.02\% for GPT-4o, and reduces hallucination to 4.76\%, compared with 10.21\% and 15.59\%, respectively. This supports the interpretation that DoRA-based adaptation improves retrieval robustness, suitable for real-world applications where retrieval potentially induces errors.

\section{Manually Curated Dataset}
\label{sec:human_annotated_dataset}

We created a high-quality QA dataset for expert-curated evaluation using a combination of LLM-driven generation, LLM judgment, and human expert annotation. This is a separate, smaller pipeline used only for the expert-curated set summarized in the main paper. We followed a five-step process that blends human expert feedback with LLM-generated data.

\paragraph{Bootstrap dataset.} We asked a domain expert to generate 25 QA pairs based on and answerable with the information in our set of domain-specific seed documents. We obtained 5 examples for each question style: \textsc{find}, \textsc{provide}, \textsc{generate}, \textsc{summarize}, and \textsc{explain}.

\paragraph{Document chunking.} We chunked the seed documents into semantically coherent passages using LumberChunker~\cite{duarte-etal-2024-lumberchunker}. We used the embedding model listed in \Cref{tab:core-model-choices} to calculate an embedding vector for each passage. This process produced 2,187 non-overlapping passages from all seed documents.

\paragraph{Semantic graph construction.} We constructed a k-NN graph (for $k=5$) with nodes corresponding to text passages and edges between two nodes representing semantic similarity estimated using cosine distance between the passages' embedding vectors. We only accepted edges between nodes when the cosine distance was less than 0.35. The resulting directed graph had 2,187 nodes and 8,121 edges with the average node degree of 3.7$\pm$0.9.

\paragraph{LLM-based dataset generation.} We prompted the expert-curated-set generator listed in \Cref{tab:core-model-choices} to generate QA pairs for each question style using context retrieved by querying the k-NN graph. First, we selected uniformly at random a node from the graph. Then, depending on the question style, we extracted the ego-graph around the selected node. For \textsc{find} and \textsc{provide} style questions, we returned only the selected node, and for the other styles we returned the one-hop neighborhood around the node, including the selected node itself. We then prompted the LLM to generate QA pairs for the given question type and context. \Cref{fig:llm_generate_qa_example_prompt} shows the prompt used for this step. The prompt includes two QA examples for in-context learning selected uniformly at random from the bootstrap dataset conditioned on the question style.

The generated question, answer, and context were then evaluated by the expert-curated-set judge listed in \Cref{tab:core-model-choices} using the prompt in \Cref{fig:llm_judge_qa}. We asked it to assign a score in the range $[0,1]$, where 0 is worst and 1 is best. We accepted generated items with scores above 0.75. This process produced 100 QA pairs with context, with 20 examples for each question style.

\paragraph{Human annotation.} Finally, domain experts were shown the generated questions and context and asked to provide an answer. We did not show the generated answer to the annotators. Of the 100 generated examples, the annotators deemed 86 valid and provided answers derived from the supplied passages. The combination of LLM-generated questions, human-generated answers, and the relevant context comprises the final dataset.

\begin{figure*}[t]
\begin{mdframed}[
  linewidth=1pt,
  linecolor=green,
  roundcorner=5pt,
  backgroundcolor=green!5
]
\begin{small}
\begin{verbatim}
You are a helpful assistant tasked with creating a single-turn question answering dataset. 
Your task is to generate one question-answer (QA) pair based on the provided text passages.

**Instructions:**
- The question must be answerable using only the information in the passages.
- The question type must match the specified type below.
- Use the passage references ([1], [2], etc.) to ground your question and answer.
- Do not use external knowledge or make assumptions beyond the passages.
- Ensure the answer is concise, factually correct, and directly supported by the text.

**Question Type:** {question_type}

**Examples:**

{examples}

**Passages:**  
{passages}

**Output Format (JSON):**
{{
    {{
      "question": "Your generated question here",
      "answer": "Your generated answer here" 
    }}
}}

\end{verbatim}
\end{small}
\end{mdframed}
\caption{Prompt template used for generating QA pairs with In-Context Learning.}
\label{fig:llm_generate_qa_example_prompt}
\end{figure*}

\begin{figure*}[t]
\begin{mdframed}[
  linewidth=1pt,
  linecolor=green,
  roundcorner=5pt,
  backgroundcolor=green!5
]
\begin{small}
\begin{verbatim}
You are an expert evaluator tasked with assessing the quality of a question-answer (QA) pair 
generated from a set of passages. Your evaluation should determine how well the QA pair aligns 
with the intended question type and whether it is accurate and contextually grounded.

Please assess the QA pair based on the following criteria:

1. **Question Type Alignment**: Does the question clearly reflect the specified question 
type (e.g., Find, Explain, Summarize)?
2. **Contextual Answerability**: Can the question be answered using only the information provided 
in the passages?
3. **Answer Accuracy**: Is the answer factually correct and directly supported by the passages?

Provide a brief reasoning for your evaluation and assign a **quality score between 0.0 (poor) 
and 1.0 (excellent)**.

**Question Type:** {question_type}

**Passages:**  
{passages}

**Question:** {question}

**Answer:** {answer}

**Output Format (JSON):**
{{
    {{
      "reasoning": "Your reasoning about the evaluation and score",
      "score": 0.0 to 1.0
    }}
}}
\end{verbatim}
\end{small}
\end{mdframed}
\caption{Prompt template used for judging the quality of generated question and answers conditioned on the question style.}
\label{fig:llm_judge_qa}
\end{figure*}

\subsection{Use as an External Evaluation Set}

The final 86 expert-answered items are used only as an external evaluation check, not as supervised fine-tuning data. This set complements the larger synthetic benchmark by testing whether the same systems remain grounded when evaluated against independently written expert answers. We evaluate models under both retrieved context and oracle gold evidence with the same semantic and RAGEval-style diagnostics used in the main experiments; results are summarized in~\Cref{tab:human-main}.

\section{Human--LLM-Judge Alignment on RAGEval-Mapped Dimensions}
\label{appd:human-judge-align}

Our main results report RAGEval Completeness, Hallucination, and Irrelevance, scored by an LLM judge following the open-source RAGEval protocol of \citet{zhu2024rageval}. To calibrate this judge against expert judgment we focus on the three rubric dimensions that correspond directly to those signals---\textit{groundedness} (hallucination counterpart), \textit{comprehensiveness} (completeness counterpart), and \textit{relevance} (irrelevance counterpart)---and score the same 50-item stratified sample described in \Cref{sec:human-eval} under the identical rubric. \Cref{tab:human-judge-align} reports the alignment between the LLM judge and the mean score across the four domain-expert annotators on these three RAGEval-mapped dimensions. The remaining four rubric dimensions (verbosity, composition, clarity, alignment) are stylistic and have no RAGEval counterpart; they are omitted from this calibration. The judge agrees substantially with the experts on \textit{groundedness} ($\kappa_w{=}0.56$, Within-1 92\%), moderately on \textit{comprehensiveness} ($\kappa_w{=}0.23$, Within-1 70\%), and very tightly within one point on \textit{relevance} (Within-1 96\%, MAD 0.19), where $\kappa_w$ is near zero because the expert scores saturate at 4--5 and leave little variance for chance correction. As noted in the main paper, we therefore treat LLM-judge metrics as scalable diagnostics for grounding-anchored signals rather than as a substitute for expert review on borderline items.

\begin{table}[t]
\centering
\footnotesize
\setlength{\tabcolsep}{2.5pt}
\renewcommand{\arraystretch}{1.05}
\resizebox{.8\columnwidth}{!}{%
\begin{tabular}{lcccc}
\toprule
\textbf{Dim.} & \textbf{Exact} & \textbf{Within-1} & \textbf{MAD} & \textbf{$\kappa_w$} \\
\midrule
Comp.  & 24.0\% & 70.0\% & 0.94 & \phantom{$-$}0.23 \\
Hall.  & 80.0\% & 92.0\% & 0.30 & \phantom{$-$}0.56 \\
Irrel. & 80.0\% & 96.0\% & 0.19 & $-$0.07 \\
\bottomrule
\end{tabular}
}
\caption{Human--LLM-judge alignment on the 50-item stratified sample. Exact, Within-1, and MAD compare the mean expert score with the LLM judge; $\kappa_w$ is quadratic-weighted Cohen's kappa.}
\label{tab:human-judge-align}
\end{table}

\section{Metrics}
\label{app:retrieval-metrics}

For each benchmark instance $i$, we treat the question $q_i$ as the retrieval query, and the gold evidence passages recorded in DoRA as the set of relevant items $G_i$ (one or more chunk IDs). A retriever produces a ranked list of retrieved passages $R_k(q_i)=(d_{i,1}, d_{i,2}, \dots, d_{i,k})$, where $d_{i,j}$ is the chunk ID at rank $j$ for query $q_i$, and $k$ is the evaluation cutoff (e.g., $k=3$ in our QA study).

\paragraph{Hit@$k$.} Let $I$ be the set of evaluated  (e.g. \ours test set). We define the instance-level indicator
\begin{equation*}
\mathrm{Hit@}k_i=\mathbf{1}\!\left[\exists\, j\in\{1,\dots,k\}:\ d_{i,j}\in G_i\right],
\end{equation*}
and report the average $\mathrm{Hit@}k=\frac{1}{|I|}\sum_{i\in I}\mathrm{Hit@}k_i$. Intuitively, $\mathrm{Hit@}k$ measures whether the retriever returns \emph{any} gold evidence chunk in its top-$k$ list for $q_i$.

\paragraph{Recall@k.} When multiple evidence passages are marked relevant, we measure coverage of the gold set:
\begin{equation*}
\mathrm{Recall@}k_i=\frac{\big|\{d_{i,1},\dots,d_{i,k}\}\cap G_i\big|}{|G_i|},
\end{equation*}
and report $\mathrm{Recall@}k=\frac{1}{|I|}\sum_{i\in I}\mathrm{Recall@}k_i$. This measures how completely the top-$k$ retrieved chunks cover the gold evidence set (with set intersection preventing repeated retrieval of the same chunk from inflating the score).

Details of retrievers are shown in~\Cref{tab:retriever-ids}.

\begin{table*}[t]
  \centering
  \small
  \setlength{\tabcolsep}{6pt}
  \renewcommand{\arraystretch}{1.1}
  \begin{tabular}{ll}
  \toprule
  Retriever & Model / Implementation ID \\
  \midrule
  BM25~\cite{robertson2009bm25} & rank-bm25  \\
  GTE-multilingual-Base~\cite{alibaba2025gte} & Alibaba-NLP/gte-multilingual-base \\
  MiniCPM-Embedding~\cite{openbmb2025minicpmembedding} & openbmb/MiniCPM-Embedding \\
  BGE-M3~\cite{xiao2024bgem3} & BAAI/bge-m3 \\
  IBM Granite (english-r2)~\cite{ibm2025graniteembeddingr2} & ibm-granite/granite-embedding-english-r2 \\
  \bottomrule
  \end{tabular}
\caption{Retriever identifiers for reproducibility.}
\label{tab:retriever-ids}
\end{table*}

\paragraph{Answer-coverage metric models.}
For reproducibility of the learned answer-coverage scores, the implementation identifier for BERTScore Recall is listed in~\Cref{tab:core-model-choices}.

\begin{table*}[t]
\centering
\scriptsize
\setlength{\tabcolsep}{3pt}
\renewcommand{\arraystretch}{1.15}
\newcolumntype{C}[1]{>{\centering\arraybackslash}p{#1}}
\newcolumntype{R}[1]{>{\RaggedRight\arraybackslash}p{#1}}
\begin{tabular}{@{} R{.22\textwidth} C{.03\textwidth} C{.03\textwidth} C{.03\textwidth} C{.03\textwidth} C{.05\textwidth} C{.03\textwidth} C{.03\textwidth} R{.44\textwidth} @{} }
\toprule
\textbf{Work} & \textbf{S} & \textbf{E} & \textbf{MH} & \textbf{MM} & \textbf{Scn} & \textbf{Sty} & \textbf{Seed} & \textbf{Notes} \\
\midrule
Few-shot Data Synthesis for Open-Domain Multi-hop QA (EACL’24)~\citep{chen2023fewshotmultihop} & \cmark & \xmark & \cmark & \xmark & Open & \xmark & \cmark & Few-shot pipeline synthesizing multi-hop QA from Wikipedia document pairs; improves smaller LMs on Hotpot‑style tasks; no dedicated RAG evaluation metrics. \\
FM2DS (EMNLP’25)~\citep{abaskohi2025fmds} & \cmark & \xmark & \cmark & \cmark & Open & \xmark & \cmark & Few-shot multimodal, multihop synthesis grounded in multimodal Wikipedia pages; introduces a multimodal multi-hop benchmark over long docs. \\
SMMQG (Findings EMNLP’24)~\citep{wu2024smmqg} & \cmark & \xmark & \xmark & \cmark & Open & \cmark & \cmark & Generates modality/style‑conditioned multimodal questions grounded in Wikipedia documents to stress MMRAG; not explicitly multi-hop. \\
Prompting‑based SDG (LREC‑COLING’24)~\citep{schmidt-etal-2024-prompting-based} & \cmark & \xmark & \xmark & \xmark & Open & \xmark & \xmark & Prompting‑based synthetic QA generation for few‑shot settings; not grounded to specific seed docs. \\
RAGAS (EACL’24)~\citep{es2023ragas} & \xmark & \cmark & \xmark & \xmark & Open & N/A & N/A & Reference‑free RAG metrics (faithfulness, answer/context relevance); framework only. \\
ARES (NAACL’24)~\citep{saad2023ares} & \xmark & \cmark & \xmark & \xmark & Open & N/A & N/A & Automated evaluation framework with faithfulness/attribution metrics and diagnostics. \\
RAGChecker (NeurIPS’24 D\&B)~\citep{ru2024ragchecker} & \xmark & \cmark & \xmark & \xmark & Open & N/A & N/A & Fine‑grained diagnostic framework for RAG (answerability, support, hallucination). \\
RAGEval (ACL’25)~\citep{zhu2024rageval} & \cmark & \cmark & \xmark & \xmark & Scn. & \xmark & \xmark & Scenario templates derived from seed docs, but QA/eval built over generated articles/configs rather than directly over the original seed docs. \\
Automatic Dataset Generation for KIQA (arXiv’25)~\citep{yuen2025automatic} & \cmark & \xmark & \xmark & \xmark & Domain & \xmark & \cmark & Generates domain QA from technical sources (e.g., manuals/specs). Emphasis on creating fine-tuning data rather than defining an evaluation framework. \\
DataMorgana (ACL’25 Ind.)~\citep{filice2025datamorgana} & \cmark & \xmark & \xmark & \xmark & Domain & \xmark & \cmark & Enterprise benchmark synthesis over private corpora with user-defined question categories to control traffic; emphasizes diversity and question--passage fit, leaving evidence-grounded answer scoring largely out of scope. \\
\midrule
\textbf{Ours (DoRA)} & \cmark & \cmark & \xmark$^{\dagger}$ & \xmark$^{\dagger}$ & Domain & \cmark & \cmark & Domain‑oriented, contamination‑aware evaluation with style‑conditioned synthetic Q\&A grounded in domain sources; task‑aware scoring and baselines. \\
\bottomrule
\end{tabular}
\caption{Comparison across synthetic data generation, RAG evaluation, realistic context dimensions, and whether question generation is grounded in seed documents. Abbrev.: S=Synthetic, E=RAG Evaluation, MH=Multi‑hop, MM=Multimodal, Scn=Scenario, Sty=Style, Seed=Seed‑doc. N/A marks evaluation frameworks rather than data‑generation methods. $^{\dagger}$Multi‑hop, Multimodal are configurable (planned stressors) rather than a core requirement.}
\label{tab:lit-compare-landscape}
\end{table*}

\section{RAGEval Metrics}
\label{app:rageval-metrics}

We compute the end-to-end faithfulness diagnostics using the open-source RAGEval evaluation implementation\footnote{\url{https://github.com/OpenBMB/RAGEval/tree/main/rageval/evaluation}} from \citet{zhu2024rageval}, with the rubric-judge setup listed in~\Cref{tab:core-model-choices}.
For each instance $i$, RAGEval extracts a set of reference keypoints $\mathcal{K}_i=\{k_{i,1},\dots,k_{i,N_i}\}$ from the gold answer $a_i$ and evaluates a system answer $\hat a_i$ by classifying each keypoint into one of three categories:
\begin{itemize}
  \item $\mathcal{K}_{i,\mathrm{comp}}$: keypoints judged as \emph{complete} (covered correctly by $\hat a_i$);
  \item $\mathcal{K}_{i,\mathrm{irr}}$: keypoints judged as \emph{irrelevant} (missing/unaddressed by $\hat a_i$); and
  \item $\mathcal{K}_{i,\mathrm{hall}}$: keypoints judged as \emph{hallucinated/wrong} (addressed incorrectly by $\hat a_i$).
\end{itemize}

\paragraph{Instance-level scores.}
The metrics are defined as normalized keypoint counts:
\begin{align*}
\mathrm{RAG\text{-}Comp}_i
  &= \frac{\lvert \mathcal{K}_{i,\mathrm{comp}}\rvert}{\lvert \mathcal{K}_i\rvert},\\
\mathrm{RAG\text{-}Hall}_i
  &= \frac{\lvert \mathcal{K}_{i,\mathrm{hall}}\rvert}{\lvert \mathcal{K}_i\rvert},\\
\mathrm{RAG\text{-}Irr}_i
  &= \frac{\lvert \mathcal{K}_{i,\mathrm{irr}}\rvert}{\lvert \mathcal{K}_i\rvert}.
\end{align*}
All three scores lie in $[0,1]$; higher completeness is better, while lower hallucination and irrelevance are better.

\paragraph{Dataset-level reporting.}
We report micro-averages over instances:
\begin{equation*}
\mathrm{RAG\text{-}Metric}=\frac{1}{|I|}\sum_{i\in I}\mathrm{RAG\text{-}Metric}_i,
\end{equation*}
where $I$ is the evaluated set (e.g., the 1,259 test split from \ours dataset).

\section{\ours Dataset}
\label{appd:dataset}

Details regarding the train/test sets used to benchmark different LLMs in the defense domain are shown in~\Cref{tab:testset,tab:trainset}.

\begin{table}[t]
  \centering
  \small
  \setlength{\tabcolsep}{6pt}
  \renewcommand{\arraystretch}{1.08}
  \resizebox{.8\linewidth}{!}{%
  \begin{tabular}{lrrr}
  \toprule
  \textbf{Task} & \textbf{Samples} & \textbf{\%} & \textbf{Docs} \\
  \midrule
  \textsc{explain}   & 1147 & 21.5\% & 20 \\
  \textsc{find}      & 986  & 18.5\% & 20 \\
  \textsc{generate}  & 1184 & 22.2\% & 20 \\
  \textsc{provide}   & 873  & 16.4\% & 20 \\
  \textsc{summarize} & 1133 & 21.3\% & 20 \\
  \midrule
  \textbf{Total} & \textbf{5{,}323} & 100.0\% & 20 \\
  \bottomrule
  \end{tabular}%
  }
  \caption{Cross-generator training-set composition by style (5{,}323-instance Claude Sonnet 4.6 training split, 90/10 train/eval), used for \ours{} SFT.}
  \label{tab:trainset}
\end{table}

\begin{table*}[t]
\centering
\small
\setlength{\tabcolsep}{5pt}
\renewcommand{\arraystretch}{1.08}
\resizebox{.8\linewidth}{!}{%
\begin{tabular}{lrrrrrrr}
\toprule
\textbf{Task} & \textbf{Samples} & \textbf{\%} & \textbf{Avg Q Len} & \textbf{Avg A Len} & \textbf{Refs/Sample} & \textbf{Duration (s)} & \textbf{Docs} \\
\midrule
\textsc{explain}   & 165 & 13.1\% & 13.67 & 33.59 & 2.99 & 14.06 & 18 \\
\textsc{find}      & 421 & 33.4\% & 12.65 & 20.83 & 1.00 &  4.79 & 20 \\
\textsc{generate}  & 309 & 24.5\% & 13.76 & 31.42 & 3.34 & 14.31 & 20 \\
\textsc{provide}   & 155 & 12.3\% & 14.00 & 30.45 & 1.91 & 12.37 & 19 \\
\textsc{summarize} & 209 & 16.6\% & 12.85 & 29.35 & 2.65 & 11.29 & 18 \\
\midrule
\textbf{Total/Avg} & \textbf{1259} & 100.0\% & 13.26 & 27.70 & 2.22 & 10.35 & $\sim$19 \\
\bottomrule
\end{tabular}%
}
\caption{Testset composition and basic statistics by style. “Refs/Sample” counts linked evidence passages.}
\label{tab:testset}
\end{table*}

\section{Example Questions and Answers by Style}
\label{appd:examples}

We use five intent styles to approximate deployment traffic in defense settings:

\begin{itemize}
    \item \textsc{find}: pinpointing exact data points (e.g., deployment dates, cost figures);
    \item \textsc{explain}: clarifying technical distinctions critical to planning and risk assessment;
    \item \textsc{summarize}: condensing lengthy reports into concise, actionable overviews;
    \item \textsc{generate}: producing structured lists of capabilities, gaps, or priorities at a glance; and
    \item \textsc{provide}: surfacing key quantitative metrics (budgets, procurement data).
\end{itemize}

Examples for each style are shown in~\Cref{tab:style-examples}.

  \begin{table*}[t]
  \centering
  \scriptsize
  \setlength{\tabcolsep}{4pt}
  \renewcommand{\arraystretch}{1.15}
  \begin{tabularx}{\textwidth}{@{} l X X @{}}
  \toprule
  \textbf{Style} & \textbf{Example question} & \textbf{Example answer} \\
  \midrule
  \textsc{find} &
  What capabilities does the Research Technology and Operations provide to Defense IS\&T? &
  Research Technology and Operations provides specialist technical and business capabilities to
  Defense IS\&T. \\
  \textsc{explain} &
  How is the concept of `asymmetric advantage' defined in a military context? &
  In a military context, `asymmetric advantage' refers to military capabilities that pit strength
  against weakness, often in a non-traditional or unconventional manner, and that disrupt a
  potential adversary's decision calculus. \\
  \textsc{summarize} &
  What initiatives are being continued and enhanced to support the growth of the defense industry
  workforce? &
  The Schools Pathways Program, Defense Industry Internship Program, and Defense Industry Pathways
  Program will continue and be enhanced to achieve better outcomes. \\
  \textsc{generate} &
  What measures will be taken to provide the industry with clearer information about Defense's
  needs? &
  The industry will receive clearer and more detailed information about Defense's needs through
  more classified briefings, the establishment of a tri-partite defense industry council, and
  industry forums. \\
  \textsc{provide} &
  What was the total value of contracts awarded by Defense in 2022--23? &
  In 2022--23, Defense awarded over \$38 billion in contracts. \\
  \bottomrule
  \end{tabularx}
  \caption{Illustrative examples of the five DoRA intent styles. Examples are shown for
  readability and to convey the expected output format.}
  \label{tab:style-examples}
  \end{table*}

\section{Detailed Methodology}
\label{appd:detailed-method}

\paragraph{Core model choices.}
\Cref{tab:core-model-choices} summarizes the default model choices by pipeline stage.

\paragraph{Computational budget and infrastructure.}
Most benchmark construction and evaluation stages use API-hosted LLMs or lightweight local encoders, so exact provider-side compute is not observable. For local components, we use the model identifiers listed in~\Cref{tab:core-model-choices}; the largest local model used during construction is Gemma-2-9B-IT for evidence scoring, run in 8-bit mode, while the main adapted generator is Llama-3.1-8B-Instruct with LoRA adapters. DoRA SFT is performed on a single NVIDIA GPU workstation; across the three fine-tuning seeds reported in~\Cref{tab:qa-rageval-gte}, the total training budget is approximately 10 GPU-hours of 1xA100 GPU (80GB). We report API model names rather than parameter counts where providers do not disclose exact deployed model sizes, and we release prompts, code, and configuration files to make the observable compute path reproducible.

\paragraph{Training setup and hyperparameters.}
For \ours{} SFT, we fine-tune Llama-3.1-8B-Instruct with LoRA adapters on the 5{,}323-instance cross-generator training split, using the 90/10 train/eval partition described in~\Cref{tab:trainset}. We use the same retrieved-context input format as the downstream QA experiments: the question is concatenated with the evidence context, and the model is trained to generate the reference answer. We tune only a small set of LoRA and optimization hyperparameters on the held-out eval split, selecting the checkpoint with the best validation loss before evaluating once on the fixed test split. The best-found configuration uses LoRA rank 64, LoRA $\alpha{=}$128, dropout 0.05,
  learning rate $1 \times 10^{-5}$, batch size 4 (effective 16 with gradient accumulation),
  gradient accumulation 4, max sequence length 4096, and 3 training epochs
  (checkpoint selected by best validation loss).
The three-seed results in~\Cref{tab:qa-rageval-gte} rerun this selected configuration with different random seeds rather than re-tuning per seed.

\begin{table*}[t]
\centering
\footnotesize
\setlength{\tabcolsep}{3pt}
\renewcommand{\arraystretch}{1.12}
\resizebox{.96\textwidth}{!}{%
\begin{tabular}{l l}
\toprule
\textbf{Stage} & \textbf{Model / setting} \\
\midrule
Q\&A generation -- test split & \texttt{openai/gpt-4o} (temp 0.3, max 512 new tokens) \\
Q\&A generation -- training split  & \texttt{anthropic/claude-sonnet-4.6} (temp 0.3, max 512 new tokens) \\
Expert-curated set: Q\&A generation & \texttt{openai/gpt-4o} \\
Expert-curated set: quality judge & \texttt{gemini-2.5-pro} \\
Expert-curated set: graph encoder & \texttt{sentence-transformers/all-MiniLM-L6-v2} \\
Human--LLM-judge calibration & \texttt{openai/gpt-4o-mini} via OpenRouter (temp 0) \\
RAGEval rubric judge & \texttt{openai/gpt-4o-mini} \\
Retrieval encoder (construction) & \texttt{sentence-transformers/all-mpnet-base-v2} \\
Retrieval encoder (retrieval-conditioned runs) & \texttt{Alibaba-NLP/gte-multilingual-base} \\
Evidence selection (NLI prefilter) & \texttt{facebook/bart-large-mnli} \\
Evidence selection (local scorer) & \texttt{google/gemma-2-9b-it} (8-bit) \\
Faithfulness: entailment (DES) & \texttt{microsoft/deberta-large-mnli} \\
Faithfulness: extractive span QA & \texttt{deepset/roberta-base-squad2} \\
Faithfulness: embedding similarity & \texttt{sentence-transformers/all-MiniLM-L6-v2} \\
Answer coverage: BERTScore Recall & \texttt{distilbert-base-uncased} \\
\bottomrule
\end{tabular}
}
\caption{Default model and metric implementation choices by stage under the cross-generator, cross-corpus design. The training and test splits use \emph{different LLM families} over \emph{disjoint} seed-document corpora.}
\label{tab:core-model-choices}
\end{table*}

\paragraph{Tokenization and chunking.}
We tokenize with \texttt{cl100k\_base}. Chunks are built by packing paragraphs to a target budget of 512 tokens (min 100; max 1024). Oversized chunks trigger a sentence-level sliding-window fallback with $\approx$25\% sentence overlap. Each chunk retains provenance (document id/name/path, page numbers, optional section header).

\paragraph{Retrieval.}
During dataset construction, candidate passage retrieval uses a hybrid dense+BM25 stage with cosine similarity (L2-normalized embeddings, inner-product search) and score mixing (0.7 dense / 0.3 BM25) before NLI-based evidence filtering. For retrieval-conditioned benchmark runs, we instead use the fixed retrievers described in the main paper (e.g., GTE-multilingual-Base or IBM Granite), with chunk size 320 tokens, overlap 64, and top-$k{=}3$.

\paragraph{Evidence selection and budgets.}
Evidence selection is conditioned on style $s$. We first retrieve candidates, then apply an NLI prefilter of size $K$, and finally use a local LLM scorer to rank candidate \emph{combinations} for completeness, complementarity, coherence, and task fit. \Cref{tab:evidence-budgets} summarizes the per-style evidence budgets and selection hyperparameters. When multi-chunk synthesis is enabled, we require at least two distinct evidence chunks for long-form styles.

\begin{center}
\centering
\footnotesize
\setlength{\tabcolsep}{2pt}
\renewcommand{\arraystretch}{1.12}
\resizebox{\columnwidth}{!}{%
\begin{tabular}{lrrrr}
\toprule
\textbf{Style} & $\mathbf{|E|}$ & \textbf{$K$} & \textbf{Comb.} & \textbf{Notes} \\
\midrule
\textsc{find}      & 1 & 20 & 6 & Extractive \\
\textsc{provide}   & 2 & 30 & 7 & Numeric content required \\
\textsc{explain}   & 4 & 60 & 8 & Long-form reasoning \\
\textsc{summarize} & 4 & 40 & 8 & Coverage-oriented \\
\textsc{generate}  & 5 & 50 & 7 & List/category synthesis \\
\bottomrule
\end{tabular}
}
\captionsetup{type=table,hypcap=false}
\caption{Evidence-selection budgets by style. $|E|$ is the target evidence-bundle size; $K$ is the NLI prefilter size; Comb.\ is the maximum number of chunk combinations evaluated.}
\label{tab:evidence-budgets}
\end{center}

\paragraph{Construction-time filtering.}
We over-generate (3$\times$ target size), remove exact duplicate questions per seed document, and apply task-aware quality control. Hard gates enforce non-empty fields and numeric content for \textsc{provide}; for multi-chunk items, accepted candidates must draw from at least two distinct evidence chunks. Evidence-grounded scoring uses NLI, span alignment, numeric consistency, and semantic relevance/coverage with the models listed above. We then stratify candidates by style and select the highest-quality items under per-style quotas.

\clearpage
\onecolumn
\section{Prompts}
\label{appd:prompts}

\subsection{Multi-Chunk Synthesis Directives}
\label{appd:prompts-multichunk}

For multi-chunk synthesis, we append the following directives (with configuration-dependent values, e.g., minimum chunks/citations) to each style prompt:

\begin{mdframed}[
  linewidth=1pt,
  linecolor=green,
  roundcorner=5pt,
  backgroundcolor=green!5
]
\begin{small}
\begin{verbatim}
MULTI-CHUNK SYNTHESIS MODE (ENFORCED):
- Integrate evidence from at least 2 distinct [Chunk] sections when they offer relevant details.
- Cite each supporting chunk explicitly by number or by quoting a short anchor
  phrase (minimum 2 citations).
- Merge overlapping insights into one cohesive answer rather than separate mini-answers.
- Reiterate key facts at the start of the answer so earlier chunks are not ignored.
- Answers lacking multi-chunk grounding will be rejected by downstream verification.
\end{verbatim}
\end{small}
\end{mdframed}

\subsection{Benchmark-Time Grounded Inference Prompt}
\label{appd:prompt-grounded-inference}

For end-to-end QA evaluation, each model receives the retrieved context (or oracle
gold evidence) and the test question using the following grounded inference
prompt:

\begin{mdframed}[
  linewidth=1pt,
  linecolor=green,
  roundcorner=5pt,
  backgroundcolor=green!5
]
\begin{small}
\begin{verbatim}
You are provided with grounded evidence. Use only this context to answer.
If the evidence is insufficient, respond with "I don't have enough context."

Context:
{context}

Question:
{question}

Answer:
\end{verbatim}
\end{small}
\end{mdframed}

\subsection{Style-Conditioned Q\&A Generation Prompts}
\label{appd:prompts-qa}

We generate each candidate instance by prompting the dataset-construction model listed in \Cref{tab:core-model-choices} with an evidence context and a style-specific template. In all prompts below, \texttt{\{context\}} is replaced with the selected evidence passages (concatenated).

\clearpage
\paragraph{\textsc{find}.}

\begin{mdframed}[
  linewidth=1pt,
  linecolor=green,
  roundcorner=5pt,
  backgroundcolor=green!5
]
\begin{small}
\begin{verbatim}
Based on the provided context, generate a factual question-answer pair that asks
for specific information (dates, numbers, names, locations, etc.) that can be
found in the text.

Context:
{context}

Requirements:
- Question should ask for specific factual information
- Answer should be a short, precise factual response directly from the text
- Copy the shortest exact phrase that answers the question (verbatim) from the supporting sentence
- If needed, include a brief quoted clause that contains the fact
- Answer must be declarative and must not be phrased as a question; do not include a question mark
- The answer must exist exactly in the provided context
- Focus on concrete details like dates, amounts, names, locations, or specific facts
- Prefer interrogatives: What/When/Where/Who/Which/How many/How much (avoid standalone 'How')
- If multiple values appear, choose the one that most directly answers the question
- Use only information present in the Context; do not add external knowledge
- Do not refer to the document/text/section; produce a self-contained Q&A

IMPORTANT - DO NOT generate questions about:
- Page numbers, indexes, or table of contents
- Who is mentioned/listed in the document
- Where information appears in the document structure
- Document metadata or navigation elements

Format your response as:
Question: [Your question here]
Answer: [Your answer here]

Example:
Question: What is the budget allocation mentioned for defense procurement?
Answer: $2.4 billion
\end{verbatim}
\end{small}
\end{mdframed}

\clearpage
\paragraph{\textsc{explain}.}

\begin{mdframed}[
  linewidth=1pt,
  linecolor=green,
  roundcorner=5pt,
  backgroundcolor=green!5
]
\begin{small}
\begin{verbatim}
Based on the provided context, generate a question that asks for an explanation
or clarification of a concept, distinction, or technical detail mentioned in
the text.

Context:
{context}

Requirements:
- Question should ask for an explanation, clarification, or distinction
  (prefer 'How…' or 'Why…' unless a clear 'What is the difference…' is present)
- Answer must be 2–3 sentences:
  • Sentence 1: direct explanation (what/how/why) using terms from the Context.
  • Sentence 2–3: tie the explanation to one or two specific supporting clauses
    (quote 2–8 words) from the Context.
- Ground the explanation in specific source sentences; avoid generic or speculative claims
- Use only information present in the Context; do not add background not present
- Where helpful, use short phrases from the source for key terms
- Do not refer to the document/text/section; produce a self-contained Q&A

Format your response as:
Question: [Your question here]
Answer: [Your answer here]

Example:
Question: What is the difference between operational capability and strategic capacity?
Answer: Operational capability refers to the immediate ability to conduct
specific military operations, while strategic capacity represents the long-term
potential to sustain and expand operations over time.
\end{verbatim}
\end{small}
\end{mdframed}

\clearpage
\paragraph{\textsc{summarize}.}

\begin{mdframed}[
  linewidth=1pt,
  linecolor=green,
  roundcorner=5pt,
  backgroundcolor=green!5
]
\begin{small}
\begin{verbatim}
Based on the provided context, generate a question that asks for a summary or
overview of the key points, and provide a concise summary as the answer.

Context:
{context}

Requirements:
- Question should ask for a summary, overview, or key points
- Answer should be a comprehensive but concise summary (2-4 sentences)
- Include the most important information from the context
- Use specific names, figures, and terms where they appear (avoid vague phrasing)
- Structure the summary logically and clearly
 - Include at least one concrete figure or named program if present
 - Cover what is being done (action), why (rationale), and scale (numbers/timeframes) when present
 - Avoid speculation or recommendations; report only what is explicitly stated
 - Use only information present in the Context; do not add external knowledge
 - Do not refer to the document/text/section; produce a self-contained Q&A

Format your response as:
Question: [Your question here]
Answer: [Your answer here]

Example:
Question: What are the key points of the cybersecurity strategy outlined in this section?
Answer: The cybersecurity strategy focuses on three main areas: strengthening
defensive capabilities through advanced threat detection systems, building
workforce expertise through specialized training programs, and enhancing
international cooperation through information sharing agreements.
The strategy emphasizes a proactive approach to cyber threats with increased
investment in both technology and human resources.
\end{verbatim}
\end{small}
\end{mdframed}

\clearpage
\paragraph{\textsc{provide}.}

\begin{mdframed}[
  linewidth=1pt,
  linecolor=green,
  roundcorner=5pt,
  backgroundcolor=green!5
]
\begin{small}
\begin{verbatim}
Based on the provided context, generate a question that asks for specific
quantitative data (budgets, numbers, percentages, dates) and provide the
precise numerical answer.

Context:
{context}

Requirements:
- Question should ask for specific quantitative information
- Answer should copy the exact numeric phrase from the context (including units)
- If the number appears within a sentence, quote the key clause containing it
- Include units where appropriate (dollars, percentages, dates, quantities)
- The answer must be factually accurate and directly from the text
- NO HEDGING: Do not use words like "likely", "may", "could",
  "approximately", "suggests". Provide definitive figures only as stated in
  the context.
- NO INFERENCE: Do not infer or estimate numbers; only copy what is explicitly present.
 - If a range is shown and the question asks for total/planned/overall, copy
   the range exactly (e.g., "$5.9bn – $7.9bn"). If the question asks for
   'approved' vs 'unapproved', pick the specific column value.
 - If no numeric value exists for your question, select a different
   quantitative question that can be answered exactly from the Context.
 - Use only information present in the Context; do not add external knowledge
 - Do not refer to the document/text/section; produce a self-contained Q&A

IMPORTANT - DO NOT generate questions about:
- Page numbers or document navigation
- Index entries or table of contents data
- Which page or section contains information
- Document structure or metadata

Format your response as:
Question: [Your question here]
Answer: [Your answer here]

Example:
Question: What is the total budget allocation for procurement mentioned in the report?
Answer: $1.2 billion
\end{verbatim}
\end{small}
\end{mdframed}

\clearpage
\paragraph{\textsc{generate}.}
\begin{mdframed}[
  linewidth=1pt,
  linecolor=green,
  roundcorner=5pt,
  backgroundcolor=green!5
]
\begin{small}
\begin{verbatim}
Based on the provided context, generate a question that asks for a structured
list (capabilities, priorities, requirements, etc.) and provide the answer as a
numbered list.

Context:
{context}

Requirements:
- Question should ask for a list of items (capabilities, priorities, gaps, requirements, etc.)
- Answer should be a numbered list with 3-6 items
- Each item must be grounded in a single supporting sentence from the Context
- Include a short quoted anchor (2–6 words) from that sentence for each item
- Do not merge information from multiple sentences into one item
- Use phrases directly from the source sentence (verbatim or near‑verbatim)
- Do not introduce items that are not explicitly stated in the Context
- Items should be substantial noun phrases (avoid weak forms like 'there is/are')
- Use only information present in the Context; do not add external knowledge
- Do not refer to the document/text/section; produce a self-contained Q&A

Format your response as:
Question: [Your question here]
Answer: [Your numbered list here]

Example:
Question: What are the key capabilities mentioned for the new defense system?
Answer: 1. Advanced threat detection using AI-powered surveillance systems
2. Rapid response coordination through integrated command networks
3. Multi-domain operational flexibility across land, sea, and air environments
4. Enhanced cybersecurity protection for critical infrastructure
\end{verbatim}
\end{small}
\end{mdframed}

\end{document}